\documentclass[lettersize,journal]{IEEEtran}
\usepackage{amsmath,amsfonts}
\usepackage{algorithmic}
\usepackage{algorithm}
\usepackage{array}
\usepackage{textcomp}
\usepackage{stfloats}
\usepackage{url}
\usepackage{verbatim}
\usepackage{graphicx}
\usepackage{cite}
\usepackage{booktabs}
\usepackage{amsmath}
\usepackage{multirow}
\usepackage[T1]{fontenc}
 
\usepackage{hyperref}

\begin{document}

\title{Video Denoising in Fluorescence Guided Surgery}

\author{
    \IEEEauthorblockN{Trevor~Seets\IEEEauthorrefmark{1} and Andreas~Velten\IEEEauthorrefmark{1}\IEEEauthorrefmark{2}}\\
    \IEEEauthorblockA{\IEEEauthorrefmark{1}Department of Electrical and Computer Engineering, University of Wisconsin-Madison, Madison, United States\\}
    \IEEEauthorblockA{\IEEEauthorrefmark{2}Department of Biostatistics and Medical Informatics, University of Wisconsin-Madison, Madison, United States\\}
    }



\maketitle

\begin{abstract}
Fluorescence guided surgery (FGS) is a promising surgical technique that gives surgeons a unique view of tissue that is used to guide their practice by delineating tissue types and diseased areas. As new fluorescent contrast agents are developed that have low fluorescent photon yields, it becomes increasingly important to develop computational models to allow FGS systems to maintain good video quality in real time environments. To further complicate this task, FGS has a difficult bias noise term from laser leakage light (LLL) that represents unfiltered excitation light that can be on the order of the fluorescent signal. Most conventional video denoising methods focus on zero mean noise, and non-causal processing, both of which are violated in FGS. Luckily in FGS, often a co-located reference video is also captured which we use to simulate the LLL and assist in the denoising processes. In this work, we propose an accurate noise simulation pipeline that includes LLL and propose three baseline deep learning based algorithms for FGS video denoising.
\end{abstract}


\begin{center}
\begin{figure*}[t]
\centering
\includegraphics[width=0.95\linewidth]{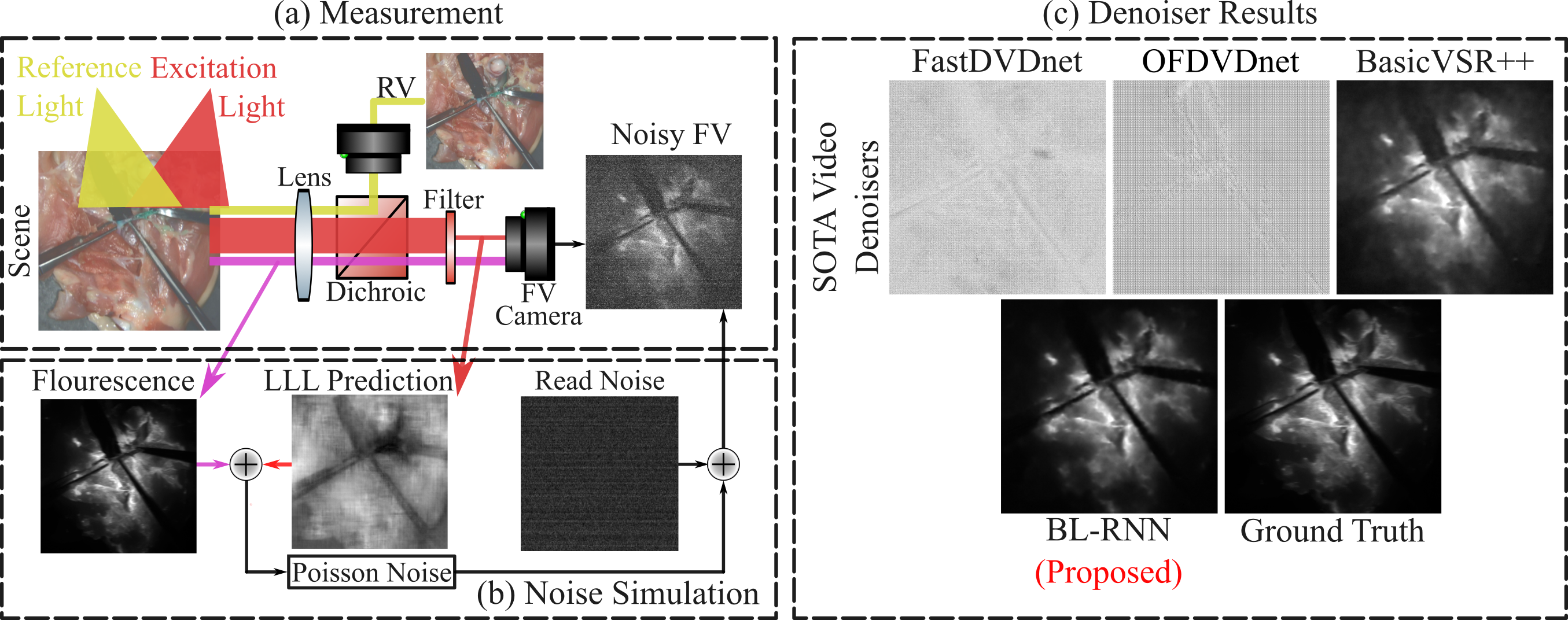}
\caption{\textbf{Measurement and Noise:} (a) In the FGS measurement process, excitation laser light and reference light shine onto the scene where the excitation light produces fluorescence at a higher wavelength. Three relevant spectral bands are imaged by the FGS system. The reference band (yellow) is isolated using a dichoric beamsplitter and imaged by the RV camera. The excitation laser band (red) is attenuated by the emission filter leaving the laser leakage light (LLL) which is combined with the fluorescence band (pink) and imaged on the FV camera.  (b) In our noise simulation, we combine a clean fluorescence image with an LLL prediction from our LLL-PN. Then we apply Poisson noise and add sampled read noise frames to produce the final noisy FV. (c) shows the results of different denoisers on a noisy FV. State-of-the-art (SOTA) video denoisers struggle while our proposed BL-RNN performs well on this task.
  \label{fig:teaser}}
\end{figure*}
\end{center}

\vspace{-30pt}
\section{Introduction}\label{sec:intro}

\IEEEPARstart{T}{here} are 320 million major surgeries performed worldwide every year in which 20 to 30$\%$ of patients require re-admittance or have serious postoperative morbidity~\cite{dobson2020trauma}. Many of these issues are due to difficulty in visualizing or identifying tissues that need removal or that should be avoided; for example, in an estimated 21$\%$ of prostate cancer removal surgeries, cancer is found on the margin of removed tissues indicating cancer was likely left in the patient~\cite{orosco2018positive}. The current standard of care (SOC) in many surgeries rely heavily on non-quantitative measures such as surgeon perception of tissue under normal lighting or tactile tissue cues. Fluorescence guided surgery (FGS) is a promising technique to improve the SOC by giving surgeons a quantifiable fluorescence video feed that helps identify the state of different tissues leading to an improvement in surgical decision making and an improvement in patient outcomes~\cite{sutton2023fluorescence}. FGS relies on a fluorescent contrast agent, either a drug or a naturally occurring fluorophore, that when imaged by an FGS imaging system helps delineate or classify tissues of interest. The most commonly used clinical contrast agents, such as  indocynanine green (ICG), are bright and operate in the near infrared while others are dim and exist in the visible light spectrum so are more challenging to capture; these dim agents may not produce enough photons at video frame rates to be clinically viable. In this work, we consider software video denoising as a relatively unexplored and promising path forward to increase the sensitivity of current systems which will increase the number of clinically viable contrast agents. 

The operation of an FGS imaging system is shown in Fig.~\ref{fig:teaser}(a); first the system emits excitation light that excites the contrast agent, then the agent emits fluorescent emission light. The system collects both emission light and reflected excitation light from a scene point then an emission filter removes much of the excitation light. However, filters are imperfect so some light is not filtered out; we call unfiltered excitation light the \textbf{Laser Leakage Light (LLL)} which can be similar in brightness to emission light, and is a core difficulty in improving FGS systems~\cite{dsouza2016review,olson2019fluorescence, pogue2023guidance}. The LLL and emission photons are added together by the fluorescence camera which outputs the \textbf{fluorescence video (FV)}. A secondary \textbf{reference video (RV)} with the same field of view is simultaneously captured by a RGB camera using spectral and temporal filtering strategies~\cite{Velten2020.08.04.236364}. The goal of a FGS denoiser is to take as input the FV and RV to produce a clean denoised FV while removing LLL. We find conventional video denoisers struggle in FGS denoising and LLL removal, so we develop a new set of baseline methods for FGS video denoising.

FGS video denoising differs from standard video denoising in four key ways. First, the RV provides a helpful source of secondary information for computer vision algorithms such as providing structural and motion cues that can be used to improve denoising performance~\cite{seets2023ofdvdnet}. We find the RV is key to simulating and removing the LL. Second, the noise levels in FGS may be much higher than in standard video denoising problems potentially requiring long range temporal integration or larger efficient models. Third, a usable FGS denoiser must be real-time capable to fit into the clinical workflow. To remain hardware agnostic, we require methods to be causal where only the past and current frames are available; most conventional video denoising methods are non-causal. 

Finally, the noise present in the FV contains the standard shot and camera noise terms, but also an additional LLL noise term. Unlike prior work~\cite{seets2023ofdvdnet} which does not consider LLL, we model the LLL term as a spatially varying shot noise term that we predict using our LLL prediction network (LLL-PN). The LLL-PN takes as input a RV frame and outputs a LLL prediction. We note our strategy for dealing with LLL may also be used to deal with noise from naturally occurring fluorescence called auto-fluorescence if it is correlated with the RV. We then use our LLL-PN within a realistic noise model, shown in Fig.~\ref{fig:teaser}(b), that accurately simulates data seen on a commercial FGS system. We use simulated data to train our denoising algorithms for a wide array of signal and noise levels before testing them on real noisy data. 

Surprisingly, we find that state of the art video denoisers when adapted and retrained for this problem struggle, as shown in Fig.~\ref{fig:teaser}(c). For example, the causal version of BasicVSR++~\cite{chan2022basicvsr++} has trouble removing the strong LLL in this example leading to signal in the denoised result where there would be none. Surprisingly, we find that NafNet~\cite{chen2022simple}, an image denoiser that uses no temporal information, outperforms these video denoisers on this dataset while the opposite is true for conventional video denoising. NafNet also is more efficient to train, taking one third of the training time as BasicVSR++. Training efficiency is extremely important when dealing with data intensive problems such as video denoising, because it  allows for practical training of larger models. Motivated by these findings we combine NafNet with different temporal propagation techniques from video denoisers to create a strong baseline model for FGS video denoising. We propose a recurrent structure, BL-RNN, with a NafNet backbone that provides robust performance with minimal network complexity and efficient training times.

\noindent\textbf{Contributions:}
\begin{itemize}
    \item We double the size of existing FGS datasets, including new data necessary for properly simulating noise and real dim signals for testing.
    \item We propose a novel method for simulating and removing spatially varying LLL in FGS, and simulate a specific commercial camera's sensor noise.
    \item We propose new network architectures for causal FGS video denoising that incorporates the most effective aspects of state of the art standard video and image denoising methods.
\end{itemize}

\begin{figure}[t!]
\centering
\includegraphics[width=0.8\linewidth]{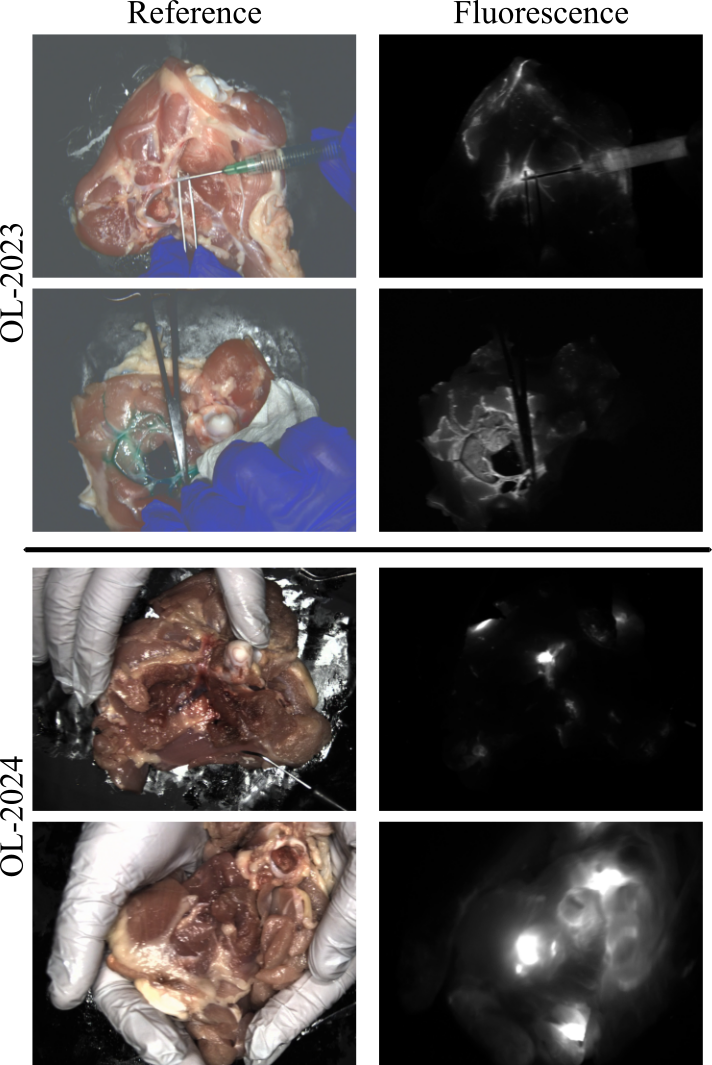}
\caption{\textbf{Dataset Example Images:} Here we show two example images for both OL-2023 and OL-2024. OL-2023 focuses on vasculature where as OL-2024 focuses on local fluorescent regions.
  \label{fig:new_dataset}}
\end{figure}

\section{Dataset}

To the best of our knowledge, the OL-2023 dataset~\cite{seets2023ofdvdnet} is currently the only publicly available FGS dataset with both FV and RV. OL-2023 contains 100 minutes of mock surgery video using the blue blood chicken surgical model \cite{chicken_thigh_model}. OL-2023 contains a number of surgical actions with a large focus on perfusion imaging with fluorescent injection sites being primarily vascular structures and focusing on slower motion scenarios. We expand the scope of this data with the new OL-2024 dataset with challenging motion scenarios and a focus on cancer and lymphatic surgeries. We follow a similar experimental setup that was used to create OL-2023;  we use the OnLume Avata clinical FGS imaging system (OnLume Surgical, Madison, WI), chicken thighs as our mock surgical patient, and high concentrations of indocyanine green (ICG) to generate very bright fluorescence that can be used as ground truth FV without LLL. In addition to the new mock surgical data we also capture a calibration and real noise test set. All videos are at 15 frames per second with a resolution of 768 by 1024 for both the RV and the FV. Dataset available at~\cite{ol24_dataset}.
\begin{figure*}[t]
\centering
\includegraphics[width=0.7\linewidth]{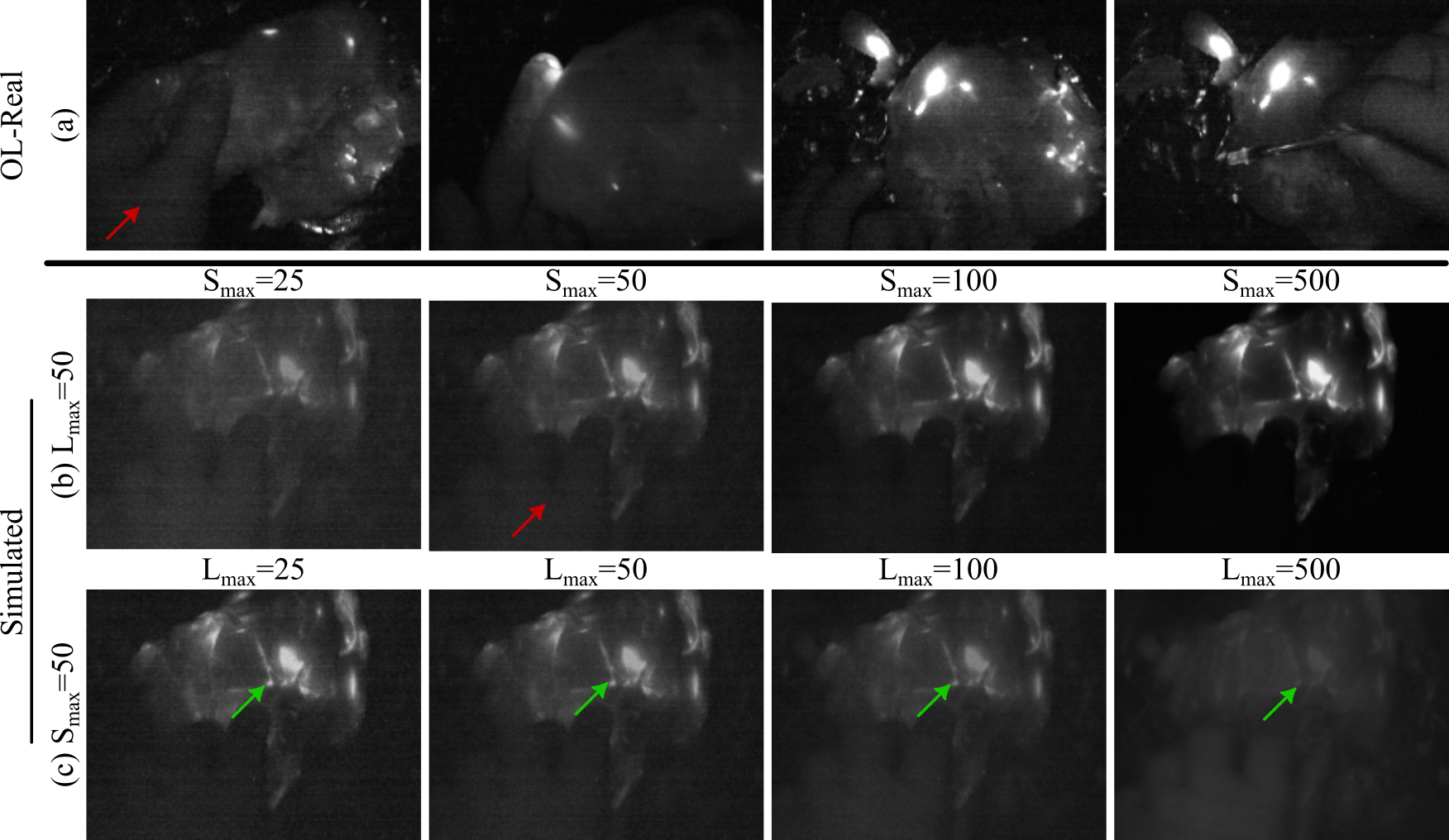}
\caption{\textbf{Real and Simulated Data:} (a) frames from our OL-Real test set. (b) simulated frames with increasing signal levels and fixed $L_{m}=50$. Qualitatively, $S_{m}=L_{m}=50$ closely matches many of the real data frames. Notice the hand (red arrow) has similar signal levels in both the real and simulated data at these parameters. (c) simulated frames with increasing $L_m$ and $S{m}=50$. Notice that the small fluorescent features (green arrow) fade as $L_m$ increases.
  \label{fig:real_vs_simulated}}
\end{figure*}

\noindent\textbf{OL-2024 and OL-Combined: } The OL-2024 dataset contains 130 minutes of new mock surgical video, each video contains aligned RV, $R^v_t$ and low-noise FV, $S_t$ frames. OL-2024 has a focus on simulating surgeries similar to cancer resections or lymphatics while also introducing more challenging motion scenarios. We primarily inject ICG into muscle or fat causing irregular fluorescent blooming with substantial scattering, modeling surgeries with patches of embedded fluorescent tissue. Example images from OL-2023 and OL-2024 are shown in Fig.~\ref{fig:new_dataset}. Additionally, unlike OL-2023, videos in OL-2024 are largely contiguous, up 18 minutes, allowing testing of long range dependencies. While these long videos are not the focus of this work we expect them to be important in future work. We combine OL-2023 and OL-2024 into core dataset, \textbf{OL-Combined} which we split into a training and testing set.

\noindent\textbf{OL-Calibration: } OL-Calibration contains three video subsets used to calibrate our realistic noise model.
\begin{itemize}
    \item \textbf{OL-Dark} contains 5,325 frames (6 min) of dark video that is captured by putting a lens cap on the system. This set is used to simulate the sensor's read noise.
    \item \textbf{OL-Phantom} contains 1,830 frames (90 seconds) of a Quel phantom (RCS-ICG-ST01-QUEL03)~\cite{ruiz2020indocyanine} for camera gain calibration. A fluorescence phantom is designed to mimic fluorescence properties of a contrast agent and acts as a fluorescent standard. The Quel phantom is a 3 by 3 grid of calibrated phantom wells of varying concentrations that is used for calibration and testing of FGS systems.
    \item \textbf{OL-LLL} contains 18,576 frames (20 min)  of a chicken thigh mock surgery~\cite{chicken_thigh_model} with no fluorescent compound, so the videos only contains ambient, laser leakage light, and read noise. We use this dataset for training our LLL-PN in our noise model.
\end{itemize}

\noindent OL-LLL and OL-Dark are split into training and testing sets.

\noindent\textbf{OL-Real:} Finally, we capture the OL-Real dataset which contains about 15 min (13,033 frames) of mock surgery with low doses of ICG that is used to test our models with real dim noisy data. We inject multiple batches of ICG with different dilutions near the noise floor of our system to produce noisy data. Example frames from OL-Real are shown in Fig.~\ref{fig:real_vs_simulated} (a).

\section{Fluorescence Noise Simulation\label{sec:noise_model}}
In order to train denoising algorithms, we add realistic noise to the OL-Combined FV frames and train our algorithms to remove this noise. We model three key sources of noise in fluorescence data, (1) shot noise originating from the inherent Poisson nature of light \cite{Hasinoff2014}, (2) camera read noise from imperfections in sensors \cite{flicker_noise,reset_noise,dark_current,fixed_pattern_noise}, and (3) laser leakage light (LLL) coming from excitation light not being fully blocked by the emission filter. The first two sources of noise (1-2) have been extensively studied for general image denoising problems \cite{noise1,wei2020physics,zhang2021rethinking} while the third source of noise is generally treated as a hardware problem and not considered in denoising contexts.

\begin{figure}[t]
\centering
\includegraphics[width=0.7\linewidth]{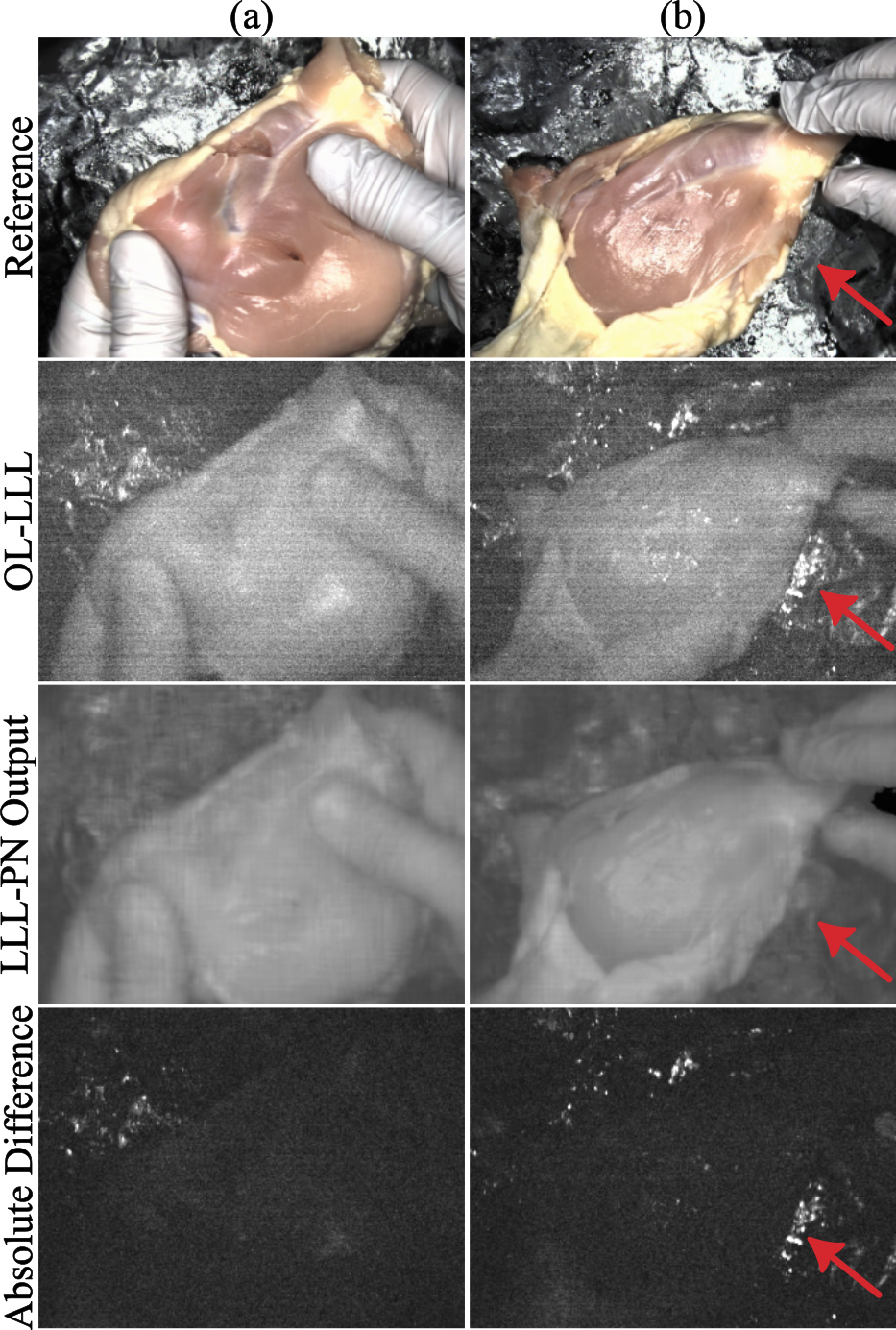}
\caption{\textbf{LLL and LLL-PN:}  This figure shows examples from OL-LLL and our corresponding LLL-PN predictions. The last row shows the absolute difference between the LLL frame and prediction. The LLL-PN is able to correctly predict most of the large structure leaving shot and read noise. The LLL-PN struggles with features that are uncorrelated to the reference image; for example, specular reflections (red arrow) are particularly difficult.
  \label{fig:LLL-PN}}
\end{figure}

We decompose the noise of our system into read noise (signal independent) and shot noise (signal dependent). We model the LLL term as a spatially varying additive photon count in the shot noise term. It is difficult to capture both the fluorescence and LLL frame simultaneously without hardware changes, so to simulate LLL we approximate it using our LLL-Prediction Network which takes the RV as input and outputs a LLL frame prediction (details in Sec.~\ref{sec:LLL_PN}). Let $S_t$ be the fluorescent signal of interest, $L^{LL}_t$ be a predicted LLL frame, and $R_t$ be the read noise of the camera at frame $t$, all scaled between 0 and 1. Then a noisy FV frame, $\Tilde{F}^v_t$, is given by,

\begin{equation}
    \Tilde{F}^v_t = \frac{1}{KS_{m}}\text{Quant}(K\text{Pois}(S_{m}S_t+L_{m}L^{LL}_t)+\frac{R_t}{R_{m}}) \label{eq:noise_model}
\end{equation}

where $\text{Pois}(\lambda)$ is a Poisson random variable with parameter $\lambda$, $K$ is the gain of the camera, Quant represents the 12-bit quantizer of the system camera, and $(S_{m}, L_{m}, R_{m})$ are scaling factors that represent the amount of noise. $S_m$ and $L_m$ represents the maximum number of fluorescence signal 
 and LLL photons in a pixel, respectively. $S_m$ and $L_m$ can vary greatly in different scenes so define the core 2 dimensional noise space we train and test on. $R_m$ controls the amount of read noise and is treated as a data augmentation parameter that will avoid overfitting on specific read noise levels that change with camera settings or ambient temperature. Note Eq.~\ref{eq:noise_model} is scaled so if there is no LLL or read noise ($L_m=0, R_t=0$), $E[\Tilde{F}^v_t] = S_t$ which we find helps in training.

In order to calibrate our noise model to a specific commercial camera we use OL-Calibration. We solve for the gain $K$ of our camera using OL-Phantom and the fact that the Poisson distribution will have mean equal to variance. We find the $K$ value that best fits a Poisson distribution over all Quel phantom wells, see supplement for details. Next we simulate $R_t$ by randomly sampling time-contiguous dark frames from OL-Dark which recent work~\cite{zhang2021rethinking} has shown to be more accurate than the common Gaussian random variable or other physics based models for read noise. Next, we use a low noise FV from OL-Combined for $S_t$, and find an estimate for $L^{LL}_t$ using our LLL-PN. In Fig.~\ref{fig:real_vs_simulated}, we show images from OL-Real as well as our simulated data with varying $S_{m}$ and $L_{m}$ levels. Notice as $L_m$ increases non-fluorescent objects begin to appear in $\Tilde{F}^v_t$; for example, hands are a good sign of significant LLL, but also generally the signal is obscured by LLL from tissue.  

\subsection{Laser Leakage Light Prediction Network\label{sec:LLL_PN}}
Our Laser Leakage Light Prediction Network (LLL-PN) takes as input a RV frame, $R^v_t$ and predicts a corresponding LLL frame, $L^{LL}_t$, to be used in our noise simulations. Our goal is to train our LLL-PN, $f_{LLL}(R^v_t)$, such that,

\begin{equation}
     f_{LLL}(R^v_t) \approx L^{LL}_t\label{eq:LLL_pn}.
\end{equation}

We choose NafNet32 for $f_{LLL}$ and train it using OL-LLL which has constant working distance and laser power which may affect the brightness of $L^{LL}_t$; these are dealt with in our noise simulation through $L_m$. For most scenes, $L^{LL}_t$ is close in intensity to the read noise of the system making it difficult to capture noise free examples. Because of this we train $f_{LLL}$ with noisy $L^{LL}_t$, similar in spirit to Noise2Noise~\cite{lehtinen2018noise2noise}. Because the RV is uncorrelated to the noise in $L^{LL}_t$, when trained with an $L_1$ loss, $f_{LLL}$ learns to predict LLL frames without shot or read noise. Specifically, an $L_1$ loss is median seeking so $f_{LLL}$ learns to predict the median value of similar patches in the noisy $L^{LL}_t$ within the training set~\cite{lehtinen2018noise2noise}. We note that our LLL-PN hinges on the assumption that the RV can be used to predict near infrared reflectivity, which we find to be true in OL-LLL, but will be important to check in human data.

Fig.~\ref{fig:LLL-PN} shows example qualitative results, note that when predicted LLL is subtracted from the noisy $L^{LL}_t$, the result still contains read noise patterns but strong LLL spatial patterns vanish. The LLL-PN struggles to predict specular reflections which is to be expected because the laser and reference lights are not co-located so specular reflections in the FV should be very difficult to predict from the RV alone without 3D scene information. We also find that other 3D dependent structures such as shadows of hands and other tools are difficult to predict for the LLL-PN. $f_{LLL}$ is able to account for $40\%$ of the total energy ($L_2$ norm)
in noisy LLL frames. This coupled with our qualitative results indicate a strong ability to predict LLL. See Suppl. Sec.~\ref{suppl_sec:lll_pn} for training details.

\section{FGS Video Denoisers\label{sec:fgs_Denoising_models}}
The goal in FGS video denoising is to take as input a set of noisy FV and RV frames, $\{\Tilde{F}^v_{\tau},R^v_{\tau}\}$, and output a set of denoised fluorescent frames $\{\hat{S}_{\tau}\}$. We simulate simulate noisy frames using the clean FV in OL-Combined as $S_t$ in our noise model and train the algorithms to recover $S_t$. Unless otherwise noted the test set used is from OL-Combined.

\subsection{Conventional denoising methods}
There are a number of candidate state of the art (SOTA) models for denoising for both image~\cite{zhang2023kbnet,chen2022simple,zamir2022restormer,zamir2021multi,chen2021hinet,wang2022uformer,tu2022maxim,cheng2021nbnet,zamir2020learning} and video~\cite{tassano2019dvdnet,BP-EVD,Tassano_2020_CVPR,wang2019edvr,yue2020supervised,qi2022real,van2021real,huang2015bidirectional,maggioni2021efficient,chan2021basicvsr,chan2022basicvsr++,cao2023learning,xiang2022remonet,chen2016deep,Arias2017VideoDV,VBM4D,hasinoff2016burst,xu2020learning,davy2018non,vaksman2021patch,seets2023ofdvdnet,tang2023automated} on conventional datasets that could be used as comparisons for our proposed baselines. However, it would be impractical and expensive to update all methods to properly use the RV, force causality, and retrain them on our dataset. Instead, we choose four models to act as our SOTA comparisons, one image denoiser and three video denoisers. We modify these four methods to incorporate the RV and to become causal. We retrain these methods on OL-Combined.

For the image denoising SOTA comparison method, we choose NAFnet~\cite{chen2022simple}. NAFnet is a state of the art image denoiser which combines key elements of other image denoisers with a goal of simplifying the core architecture resulting in a light weight network with strong performance. NAFnet uses a simplified channel attention~\cite{hu2018squeeze} for global information aggregation, depth wise convolutions~\cite{han2021demystifying,liu2022convnet} for local information, layer normalization~\cite{ba2016layer} to stabilize training and "Simple Gate" as the non-linearity. 

We split the video denoisers into three categories based on their temporal propagation strategy and choose one method from each category as our SOTA comparison method. The first category is \textbf{sliding window} strategies~\cite{tassano2019dvdnet,BP-EVD,Tassano_2020_CVPR,wang2019edvr,yue2020supervised,qi2022real,van2021real} which have a fixed temporal receptive field that is sequentially moved throughout the video sequence. We choose FastDVDNet~\cite{Tassano_2020_CVPR} due to its strong conventional video denoising performance. FastDVDNet relies on a cascaded approach where successive frames are grouped together as inputs into an initial U-net~\cite{unet} block whose outputs are fused with temporally adjacent blocks as input into a second U-net. 

The second strategy is based on \textbf{Recurrent Neural Networks} (RNNs)~\cite{huang2015bidirectional,maggioni2021efficient,chan2021basicvsr,chan2022basicvsr++,cao2023learning,xiang2022remonet,chen2016deep}. RNNs can pass information forward and backward in time recursively throughout the video, in principle allowing for long range temporal aggregation. We choose BasicVSR++~\cite{chan2022basicvsr++} because it achieves SOTA conventional denoising performance~\cite{chan2022generalization}. BasicVSR++ uses a grid propagation strategy based on flow guided~\cite{chan2021understanding} deformable convolutions~\cite{dai2017deformable,zhu2019deformable} with second order connections where features are aligned, aggregated and propagated in time.

The third temporal strategy are \textbf{Align and Merge} (A\&M) methods~\cite{Arias2017VideoDV,VBM4D,hasinoff2016burst,xu2020learning,davy2018non,vaksman2021patch,seets2023ofdvdnet,tang2023automated}. A\&M strategies in deep learning are extensions of block~\cite{Arias2017VideoDV,VBM4D} or patch matching~\cite{hasinoff2016burst} non-learning based methods. These strategies rely on combining multiple observations of the same object across time to reduce noise variance. These methods align or match regions of the video to deal with motion and then combine these matched regions in a merge step to denoise the video. For this category we choose OFDVDnet~\cite{seets2023ofdvdnet} because it was developed for FGS denoising. OFDVDnet uses an explicit align and merge step to aggregate many frames of temporal information to be used as input into downstream CNN without increasing the memory requirements of training large temporal receptive fields. 

\subsubsection{Modifications to SOTA Methods} We modify the SOTA methods to include the RV and become causal.

\noindent\textbf{RV incorporation: } We incorporate the RV in two locations for the SOTA comparison methods. First, we append it in the channel dimension on the input layer allowing the network to fully use the RV information. Second, when possible we use the RV in any relevant alignment module. Using the RV for alignment has been shown to produce better results~\cite{seets2023ofdvdnet} because it has less noise than the FV. 

\noindent\textbf{Causality: } All SOTA methods are non-causal so need to be modified. We do this by ensuring no future frames can be accessed by the present frame. In FastDVDnet and OFDVDnet we do this by shifting the cascaded U-nets to take in the 4 previous frames along with the current frame instead of using the 2 previous and 2 future frames. BasicVSR++ uses a bi-directional recursive propagation strategy with alternating branches moving information forward or backward in time. In BasicVSR++ we ensure causality by only using switching the direction of the backward propagation branches to forward propagation branches while maintaining the same network size. We indicate these are the causal versions of these SOTA networks with the superscript $^C$, for example BasicVSR++$^C$.

\subsubsection{SOTA Results} We find that the video denoising SOTA methods are unsuitable for this problem. FastDVDnet$^C$ and OFDVDnet$^C$ are unable to deal with the LLL leading to poor performance. While BasicVSR++$^C$ is able to produce reasonable results, it occasionally struggles with large motion and LLL removal. Table.~\ref{tab:sota_results} displays the PSNR/SSIM the SOTA comparison methods at the test values of $S_m = L_m = 50$. FastDVDnet$^C$ and OFDVDnet$^C$ are unable to produce reasonable results at all with a PSNR under 7dB. NafNet outperforms BasicVSR++$^C$ by 4.3dB while also using one third of the training resources. This is a surprising result, so we test whether or not this is a property of our data or the models by training our causal version of BasicVSR++$^C$ and NafNet32 on Davis~\cite{davis_dataset} which is commonly used in video denoising. BasicVSR++$^C$ outperforms NafNet32 by a large margin at all noise levels on Davis (details in Suppl. Sec.~\ref{suppl:davis_results}), indicating that NafNet32 has unique properties suited to denoising FGS data. This finding motivates the design of our baseline models.

\begin{table}[t]
    \centering
    \caption{\textbf{SOTA Results: } State of the art video denoising models retrained on OL-Combined struggle to outperform NafNet32, an image denoiser. Metrics are listed as PSNR/SSIM and a higher value is better for both. Results are for $S_m = L_m = 50$.  }
    \begin{tabular}{c|c}
         Model&  PSNR/SSIM\\ \hline
         NafNet32& 34.8/0.735\\
         BasicVSR++$^C$ & 30.5/0.639\\
         OFDVDnet$^C$ & 2.5/0.000\\
         FastDVDnet$^C$ & 6.1/0.020\\
         \hline
    \end{tabular}
    
    \label{tab:sota_results}
\end{table}

 \begin{figure*}[t]
\centering
\includegraphics[width=0.85\linewidth]{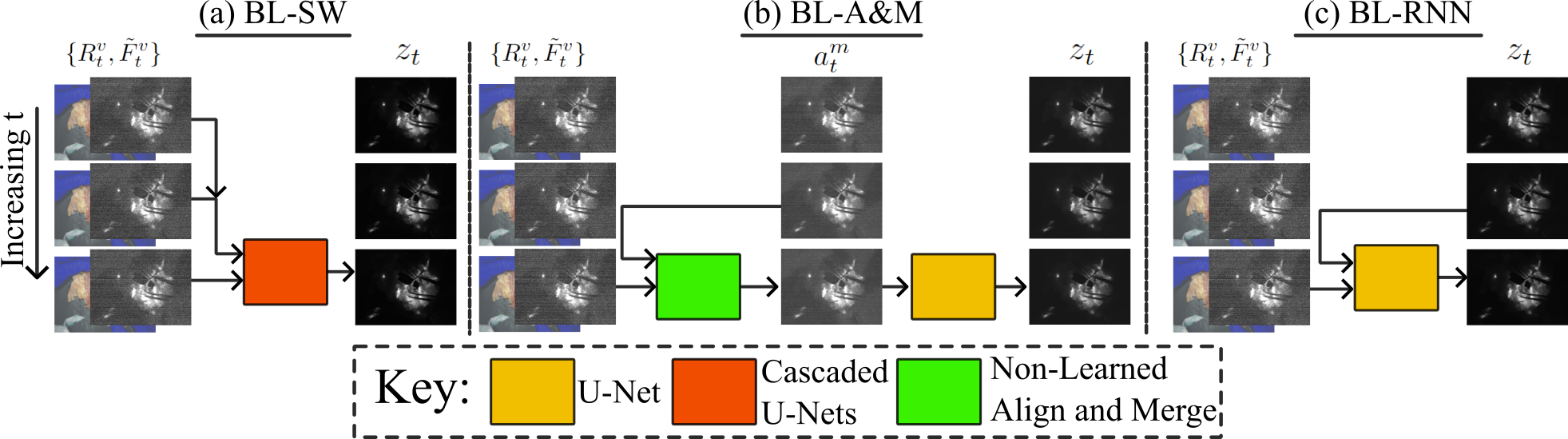}
\caption{\textbf{Baseline Model Overview:} We propose 3 baseline: (a) SW-BL uses a fixed temporal receptive field, (b) BL-A$\&$M uses a recursive non-learning based align and merge module followed by a U-net block, and (c) RNN-BL uses a simple recursive input.
  \label{fig:baseline_overview}}
\end{figure*}

\subsection{FGS Baseline Models}
Because NafNet shows such strong performance on OL-Combined, we choose to combine the core components of Nafnet with temporal propagation ideas of each temporal category to create a set of new baseline models. We choose to create simple baselines when possible because we believe strong simple baselines are important for future comparisons. An overview of our baseline models is shown in Fig.~\ref{fig:baseline_overview}. 

\noindent\textbf{Residual Blocks: } For the primary residual blocks in our baseline models we use the NAFblock~\cite{chen2022simple} from NAFnet which contains a channel attention module for global information, layer norms for stable training, and depth wise convolutions to lower parameters in the convolutional layers. We use the popular U-net structure as the primary backbone of all models, represented as $f(\cdot;\Theta)$ in the following subsections. 

\subsubsection{BL-SW}
A sliding window strategy takes as input a fixed number of previous frames and the current frame to denoise the current frame. We use a cascaded U-net structure similar to FastDVDnet. Our sliding window baseline (BL-SW) takes as input 5 frames,

\begin{align}
    \hat{S}^k_t &= f_1(\{\Tilde{F}^v_{\tau},R^v_{\tau}\}_{\tau=t-k-2}^{t-k} ;\Theta_1) \\
    \hat{S}_t &= f_2(\{\hat{S}^k_t\}_{k=0}^2,\{R^v_{\tau}\}_{\tau=t-4}^{t} ;\Theta_2).
\end{align}
where $\hat{S}_t$ is the denoised output, and $\hat{S}^k_t$ are intermediate feature frames with double the number of channels as the input. $f_1$ and $f_2$ are NafNet24 U-nets. Note FastDVDnet only computes $\hat{S}^0_t$ and $\hat{S}^2_t$, so BL-SW has an extra U-Net forward pass. But $\hat{S}^k_t$'s can be cached from previous $t$'s, so BL-SW does not require much more computation.

\subsubsection{BL-A$\&$M}
The baseline align and merge (BL-A$\&$M) first uses a non-learned function that temporally aligns and then merges frames together before the output is fed into a downstream network. While there are many candidates for both the alignment and merge functions, we choose to use the A$\&$M strategy from OFDVDnet~\cite{seets2023ofdvdnet} because it is the SOTA in FGS denoising. The causal version of the A$\&$M strategy in OFDVDnet, $A^M$, relies on calculating optical flow between every successive frame, and using a recursively defined motion corrected sum with occlusion rejection to aggregate information across time. Our BL-A$\&$M follows the following formulation,

\begin{align}
     a^m_t &= A^M(\{\Tilde{F}^v_{\tau},R^v_{\tau}\}_{\tau=t-1}^{t},a^m_{t-1}) \\
     \hat{S}_t &= f(a^m_t,\Tilde{F}^v_t, \Tilde{R}^v_{t}; \Theta).
\end{align}

Where $a^m_t$ is the intermediate temporally averaged frame, equivalent to OFDV-Forward from~\cite{seets2023ofdvdnet}. We change the downstream network, $f(\cdot;\Theta)$, used in OFDVDnet to match Nafnet32.

\subsubsection{BL-RNN}
A recurrent strategy uses past computation in a recurrent manner, for this baseline model we use a simple recurrence relation that uses a previous denoised output frame as input into the next denoising block. Our baseline recurrent neural network (BL-RNN) takes the following form,
\begin{equation}
    \hat{S}_t = f(\hat{S}_{t-1},\{\Tilde{F}^v_{\tau},R^v_{\tau}\}_{\tau=t-1}^{t}; \Theta).
\end{equation}

where $\hat{S}_t$ is the output frame at time $t$ and $f$ is a NafNet32 U-net. We found that for this model replacing the SimpleGate activation functions with ReLu activation functions provide slight performance improvements when LLL is present, so we use ReLu activation functions in our BL-RNN.  We experimented with using the recurrent structure from BasicVSR++ with NafNet residual blocks, but found that these networks where unstable and produced large artifacts.

\subsection{Training Details: }
 All networks are trained with a cosine annealing training scheme with the Adam optimizer~\cite{Kingma2014AdamAM}, and Charbonnier loss~\cite{CharbonnierLoss} because it stabilizes training by reducing impact of outliers~\cite{lai2017deep}. We train the networks for a maximum of 2 weeks on the same hardware (Nvidia A100), we choose 2 weeks as this is how long it takes to train BasicVSR++$^C$ following the original training scheme (300k iterations). We find FastDVDnet$^C$, NafNet32, and BL-RNN converge in only 1 week, so we stop training at 1 week for these models to prevent over-fitting. In order to speed up training of the A$\&$M model, we first train the NafNet32 backbone for 300k iteration before training the full model for an additional 100k iterations. In order to augment our data and allow for changes in imaging distance, laser power, and other changes to the physical system, we randomly sample from a variety of $S_{m}$, $K$, $L_m$, and $R_m$ parameters in our noise simulation during training. We train over $S_m=[10,\frac{1}{2K}]$, $L_m=[0,S_m]$, $\frac{1}{K}=[1200,2400]$, and $R_m=[4,8]$. More specifics on data augmentation and training parameters are available in the supplement and detailed configurations will be available with the code.

\begin{figure*}[t]
\centering
\includegraphics[width=0.9\linewidth]{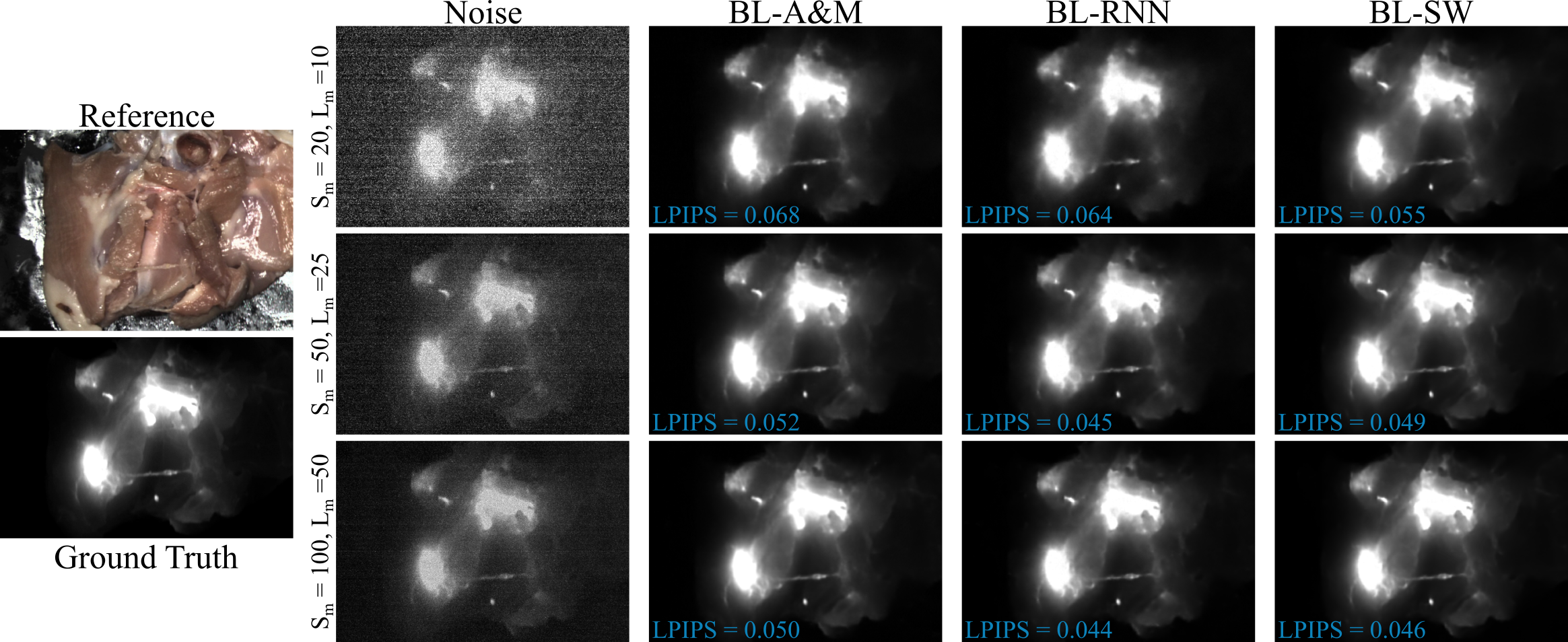}
\caption{\textbf{Proposed Baseline Results: } This figure shows the results of our baseline models over different numbers of photons with constant $\frac{L_m}{S_m}$. Notice as signal gets lower the read noise structure becomes more significant and BL-SW outperforms, whereas at higher signal levels BL-RNN gives the best results.
  \label{fig:changing_signal}}
\end{figure*}


\begin{table*}[b]
\centering
\caption{\textbf{Metric Performance: } These tables show our test metrics at (top) $S_m=50$ with changing $L_m$ values, and (bottom) at $L_m=\frac{1}{2}S_m$ with changing $S_m$ values. The arrow next to the metrics indicate the direction of better quality. Best results are \textbf{bold}.}
\begin{tabular}{lccccc|c}
\toprule
& \multicolumn{6}{c}{PSNR$\uparrow$ / SSIM$\uparrow$/ LPIPS$\downarrow$ at $S_m=50$}  \\
    \cmidrule(lr){2-7} 
Model & $L_m=0$ &$L_m=12.5$  & $L_m=25$& $L_m=37.5$& $L_m=50$ & Avg \\ \hline
BasicVSR++$^C$ & 37.15/0.83/0.108          & 35.98/0.79/0.108                    & 34.42/0.74/0.119 & 32.67/0.68/0.159 & 30.51/0.64/0.221 & 34.14/0.74/0.143 \\ 
NafNet32 & 43.06/0.95/0.063                & 40.68/0.89/0.070                     & 38.26/0.82/0.079 & 36.42/0.77/0.094 & 34.78/0.74/0.113  & 38.64/0.83/0.084 \\ 
BL-SW  & \textbf{43.50}/\textbf{0.94}/0.058 & \textbf{41.42}/0.89/0.060           & 39.54/0.83/0.065 & 37.94/0.80/0.076 & \textbf{36.44}/0.77/0.092   &  \textbf{39.77}/0.84/0.070\\ 
BL-A$\&$M &  38.59/0.89/0.063              & 38.15/0.88/0.063                   & 37.55/0.87/0.064 & 36.84/\textbf{0.86}/\textbf{0.065} & 35.92/\textbf{0.85}/\textbf{0.068} &  37.41/\textbf{0.87}/0.065\\ 
 BL-RNN &   41.67/0.92/\textbf{0.053}      & 41.11/\textbf{0.91}/\textbf{0.053} & \textbf{39.88}/\textbf{0.89}/\textbf{0.056} & \textbf{37.96}/0.84/\textbf{0.065} & 36.05/0.79/0.084 &  39.33/\textbf{0.87}/\textbf{0.062} \\ 
 \midrule\midrule
 & \multicolumn{6}{c}{PSNR$\uparrow$ / SSIM$\uparrow$/ LPIPS$\downarrow$ at $L_m=\frac{1}{2}S_m$}  \\
    \cmidrule(lr){2-7} 
Model & $S_m=5$ & $S_m=10$ & $S_m=25$ & $S_m=50$ & $S_m=100$& Avg \\ \hline
BasicVSR++$^C$ & 27.22/0.30/0.571 & 29.80/0.50/0.430 & 32.66/0.67/0.223 & 34.40/0.74/0.119 & 34.53/0.73/0.119 & 31.72/0.59/0.293 \\ 
NafNet32       & 33.87/\textbf{0.84}/0.133 & 36.09/0.86/0.088 & 37.74/0.85/0.080 & 38.25/0.82/0.079 & 38.73/0.80/0.079 & 36.94/0.83/0.092 \\ 
BL-SW          & \textbf{34.33}/0.80/\textbf{0.102} & \textbf{36.41}/0.82/\textbf{0.078} & \textbf{38.44}/0.83/0.069 & 39.54/0.83/0.065 & 40.33/0.84/0.062 & \textbf{37.81}/0.82/\textbf{0.075} \\ 
BL-A$\&$M      & 31.56/0.70/0.242 & 34.12/0.79/0.137 & 36.34/0.85/0.079 & 37.54/0.87/0.064 & 38.24/\textbf{0.89}/0.059 & 35.56/0.82/0.116 \\ 
BL-RNN         & 32.05/0.75/0.241 & 35.07/\textbf{0.83}/0.128 & 38.35/\textbf{0.88}/\textbf{0.067} & \textbf{39.97}/\textbf{0.89}/\textbf{0.055} & \textbf{40.71}/0.87/\textbf{0.054} & 37.23/\textbf{0.84}/0.109 \\ \bottomrule
\end{tabular}
\label{tab:large_table_results}
\end{table*}


\section{Results and Discussion}

\subsection{Changing $S_m$ and $L_m$}
Our noise model is parameterized by two parameters, $S_m$ and $L_m$ which represent the levels of signal and LLL, respectively. This two dimensional noise space leads to difficulty in finding the best model, and we find our baseline models are stronger in different noise regions than others. We evaluate our models throughout a range of $S_m$ and $L_m$ values, we evaluate our models with 3 metrics: PSNR as a per-pixel measure, SSIM~\cite{wang2004image} as a perceptual metric, and LPIPS~\cite{zhang2018unreasonable} as a deep learning based perceptual metric which is trained to match human perception. In general, we found LPIPS to most agree with our perceptual ranking of images where PSNR favored over-smoothing which is problematic for small features.

First, we test our models performance for a changing $L_m$ at a fixed $S_m=50$ as well as a changing $S_m$ for a fixed $\frac{L_m}{S_m}$ ratio, our quantitative metrics are summarized in Table~\ref{tab:large_table_results}. In general we find that our baseline models all perform reasonably well, outperforming the SOTA comparison models; however, no model is best for all noise parameters. For example Fig.~\ref{fig:changing_signal} shows an example where our baseline models provide reasonable performance with BL-RNN providing the best LPIPS at high photon counts, but at lower photon counts BL-SW becomes the best performing model. Also notice that in this example the difference between qualitative performance between $S_m=100$ and $S_m=50$ for all 3 baselines is very small, this agrees with the same columns in Table.~\ref{tab:large_table_results} which shows only small changes in LPIPS for the three baseline models. While more photons improve performance, this trend shows there may be a point of diminishing returns when designing drug, device, and denoising model combinations where more photons do not provide large benefits.

\begin{figure}[t!]
\centering
\includegraphics[width=0.7\linewidth]{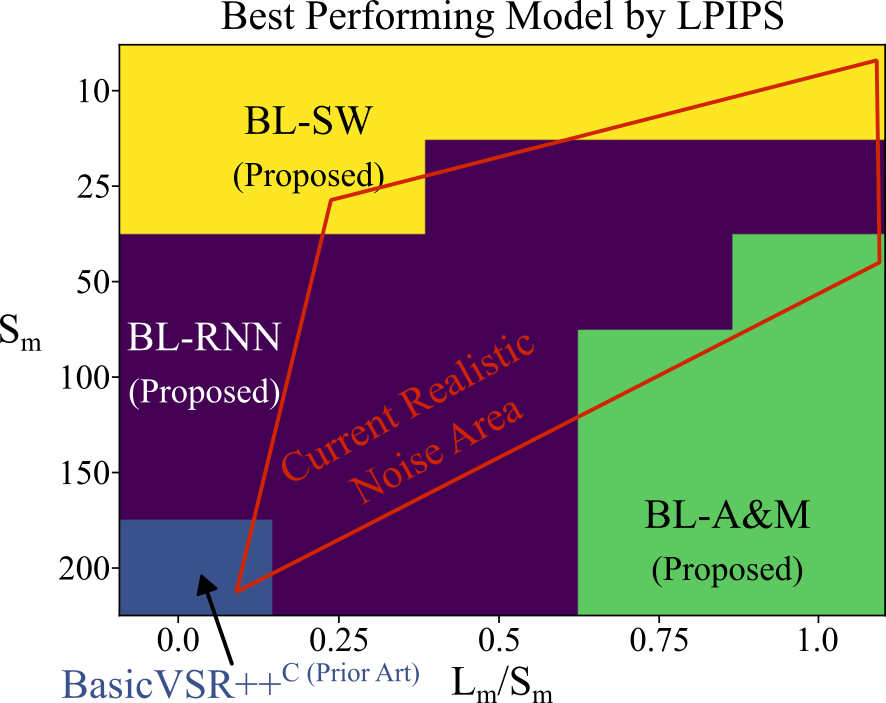}
\caption{\textbf{Best Model over $S_m$ and $L_m$:} This figure shows the model with the lowest LPIPS over a range of different $S_m$ and $L_m$ values. Inside of the red polygon indicates an estimate of realistic noise scenarios that may be seen on current systems in scenes that require denoising. 
  \label{fig:best_lpips}}
\end{figure}

Fig.~\ref{fig:best_lpips} provides a summary image showing which model obtains the best LPIPS over a larger range of noise values. We test our models for $S_m=[10,25,50,100,150,200]$ and $\frac{L_m}{S_m}=[0,0.25,0.5,0.75,1]$. BL-SW performs the best at low photon levels where read noise is significant, possibly because BL-SW has access to multiple read noise measurements that are temporally correlated allowing it to better remove the read noise. BL-A$\&$M performs well at high signal and high LLL indicating its robustness to LLL and ability to properly remove it. BL-RNN provides strong performance in the middle noise regions that most closely align with where we expect current FGS systems to operate within for noisy scenes. We find generally, BL-RNN does not over-smooth like the other baselines, likely because it keeps high frequency information to pass on in the recurrent connections. This lack of over-smoothing explains BL-RNN's superior LPIPS score in the middle noise regions. Interestingly, BasicVSR++$^C$ performs quite well at the test point $S_m=200, L_m=0$ which corresponds closely to the noise levels found in general video denoising settings, which reinforces our finding that the changes to the noise in FGS video denoising has strong impacts on performance of models developed solely for conventional video denoising. 

\begin{figure*}[t!]
\centering
\includegraphics[width=0.7\linewidth]{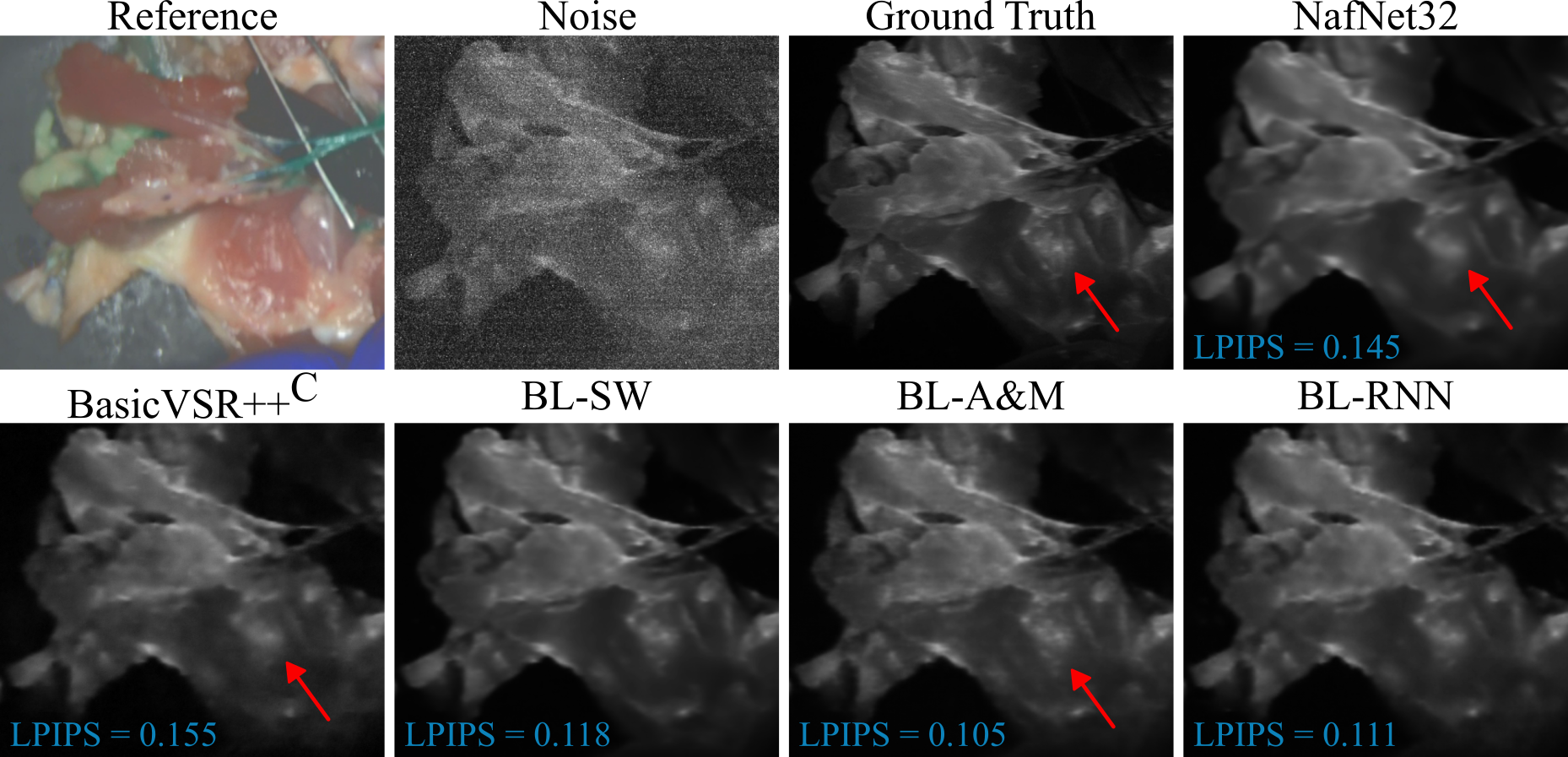}
\caption{\textbf{Example Denoised Results:} This figure shows a test case with $S_m=L_m=50$. In this example our three baseline models all perform reasonable well whereas the comparison models tend to over-smooth (red arrow).
  \label{fig:sm50_lm50_vid141}}
\end{figure*}

Fig.~\ref{fig:sm50_lm50_vid141} shows an example result where the comparison models of NafNet32 and BasicVSR++$^C$ tend to oversmooth whereas our baseline models all perform qualitatively very similar and retain small features. Generally, NafNet32 and BasicVSR++$^C$ tend to oversmooth and produce flickering videos. 

\begin{table*}[htbp]
  \centering
  \caption{\textbf{LLL Robustness: } This table shows the results of our LLL robustness tests for 2 different LLL-PNs. The models are trained with $f_{lll}^1$ but tested with either $f_{lll}^1$ or $f_{lll}^2$. We report the average PSNR, average LPIPS and our robustness measure, $\mathbf{m}_{lll}$ (closer to 0 is better). Best results are bold.}
    \begin{tabular}{lcccccc}
    \toprule
     LLL-PN& \multicolumn{3}{c}{\textbf{$f_{lll}^1$}} & \multicolumn{3}{c}{\textbf{$f_{lll}^2$}} \\
    \cmidrule(lr){1-4} \cmidrule(lr){5-7}
    \textbf{Model} & $\mathbf{m}_{lll}\uparrow$ & $\Bar{\textbf{PSNR}}\uparrow$ &$\Bar{\textbf{LPIPS}}\downarrow$& $\mathbf{m}_{lll}\uparrow$  & $\Bar{\textbf{PSNR}}\uparrow$&$\Bar{\textbf{LPIPS}}\downarrow$ \\
    \midrule
    BasicVSR++$^C$ & -6.64 & 34.14 &0.14 & -7.25 & 33.55&0.14 \\
    NafNet32 & -8.33 & 38.64 & 0.084&-12.74 & 36.32&0.091 \\
    BL-SW & -7.04 & \textbf{39.77}&0.070 & -11.28 & 37.48&0.070 \\
    BL-A$\&$M & \textbf{-2.66} & 37.41&0.65 & \textbf{-4.42} & 36.67&0.067 \\
    BL-RNN & -5.76 & 39.33 &\textbf{0.062}& -7.83 & \textbf{38.28} & \textbf{0.059}\\
    \bottomrule
    \end{tabular}
  \label{tab:LLL}
\end{table*}

\subsection{Robustness to LLL}
LLL Robustness is an important property of a strong FGS denoiser for generalizing to real data. We propose to measure LLL robustness through the change in performance as $L_m$ increases, as well as in a new test case where the LLL-PN changes to an new LLL-PN not seen during training. 

\noindent\textbf{Measuring Robustness:} We experimentally find that for a fixed $S_m$ the PSNR is roughly linear with $L_m$. This inspires us to propose using the slope of this line, $\mathbf{m}_{lll}$, in order to measure robustness to LLL levels. A lower value of $\mathbf{m}_{lll}$ indicates that the models performance is robust to changing LLL because it has a small expected change with respect the magnitude of the LLL. Specifically, we assume a model's test PSNR at a fixed $S_m$ can be written as a line,

\begin{equation}
    \text{PSNR} \approx \mathbf{m}_{lll} \frac{L_m}{S_m} + b_{lll}
\end{equation}

where $b_{lll}$ is the intercept and $\mathbf{m}_{lll}$ is the slope of the line. We find $\mathbf{m}_{lll},b_{lll}$ values with least squares regression. 

\noindent\textbf{Switching LLL-PN:} It is important for a denoising model to be robust to inaccuracies in the LLL-PN used at training time, $f_{lll}^1$; if $f_{lll}^1$ is not accurate to the true LLL in a scene it will be important for the denoising model to still perform well. We test the robustness of these models by creating a second LLL-PN, $f_{lll}^2$, by switching the training and testing data used to train $f_{lll}^1$. We use $f_{lll}^2$ to test the performance of these models under a different LLL-PN than the models were trained on. 

We test results for the parameters $S_m=50$ with $L_m=[0,12.5,25,37.5,50]$ and for [$f_{lll}^1$,$f_{lll}^2$]. We report the mean PSNR, and $\mathbf{m}_{lll}$ over this range in Table~\ref{tab:LLL}. The $R^2$ value for all linear fits for finding $\mathbf{m}_{lll}$ are above $0.95$ indicating a good fit. All models exhibit worse performance under $f_{lll}^2$ indicating a more accurate LLL-PN will be helpful in increasing generalizability of the models. The BL-RNN and BL-A$\&$M changed the least from testing with $f_{lll}^1$ to $f_{lll}^2$ and both had small $\mathbf{m}_{lll}$ indicating these models are the most robust to LLL changes. With BL-A$\&$M providing more robustness at the cost of slightly worse performance when compared to BL-RNN. Likely, BL-A$\&$M and BL-RNN are robust because their inputs contain some processed version of the FV. In the case of BL-A$\&$M the input contains a temporally averaged image so in principle contains averaged results of the LLL potentially reducing the ability for the model to overfit to a specific LLL-PN. Similarly, the BL-RNN uses a denoised output as input allowing it to gain information on the LLL from the previous frame rather than relying solely on the RV and FV, allowing for a multistage approach in estimating the LLL.

\subsection{What component allows for LLL removal?}
The SOTA models for standard video denoising have difficulty in removing the LLL. Here we do an ablation study on our BL-RNN model to test what component is important for LLL removal. We test the following:
\begin{itemize}
    \item  We remove the channel attention modules to test the importance of global information.
    \item We change the Relu activations to SimpleGate.
    \item We remove the layer norms.
    \item We remove the reference video.
\end{itemize}
We retrain each case using the same training scheme as our full network. We test at $S_{m}=50$ and $L_{m}=[0,25,50]$; we report the PSNR difference between the full BL-RNN and the ablation models in Tab.~\ref{tab:LLL_ablation}. We find that the layer norms are the most important aspect of the network and are crucial for stable training, without layer norms we needed to lower the learning rate of the network by a factor of ten for stability, but the network still could not produce strong performance at convergence. We found that the simplified channel attention module was crucial for denoising at higher $L_m$ indicating that global information is very important for removing the LLL leading to a 3.7dB improvement in PSNR. Similarly, the RV was helpful in removing LLL; although, surprisingly, when there is no LLL the network without no RV performs slightly better but the stark performance drop at higher $L_m$ makes the RV necessary. This behavior could indicate that requiring networks to denoise over a large range of $L_m$ values is difficult and without the RV this network can focus more on denoising the scenario without LLL because its performance in significant LLL will be bounded by its ability to remove it which is challenging without the RV. We also found a slight dip in LLL rmeoval when using SimpleGates instead of ReLus although other components had a larger effect.  

\begin{table}[]
    \centering
    \caption{\textbf{BL-RNN Ablation Results: } This table shows the PSNR change of our BL-RNN model with different components removed or changed. We test the models at $S_m=50$ and at three $L_m$'s. }
    \label{tab:LLL_ablation}
    \begin{tabular}{lccc}
    \toprule
    &\multicolumn{3}{c}{$L_m$}\\
   \cmidrule(lr){2-4}
       \textbf{Ablations}  &  0 & 25 & 50\\
       \midrule
        No Layer Norm     & -22.3 & -20.5 & -16.7\\
        ReLu $\rightarrow$ SimpleGate    &0.069  & -0.464&-0.401\\
        No RV    & 0.545&-1.992 &-1.054\\
        No Channel Attention & -0.553 &-1.505 &-3.708\\ \bottomrule
    \end{tabular}
    
\end{table}

\begin{figure}[t]
\centering
\includegraphics[width=0.8\linewidth]{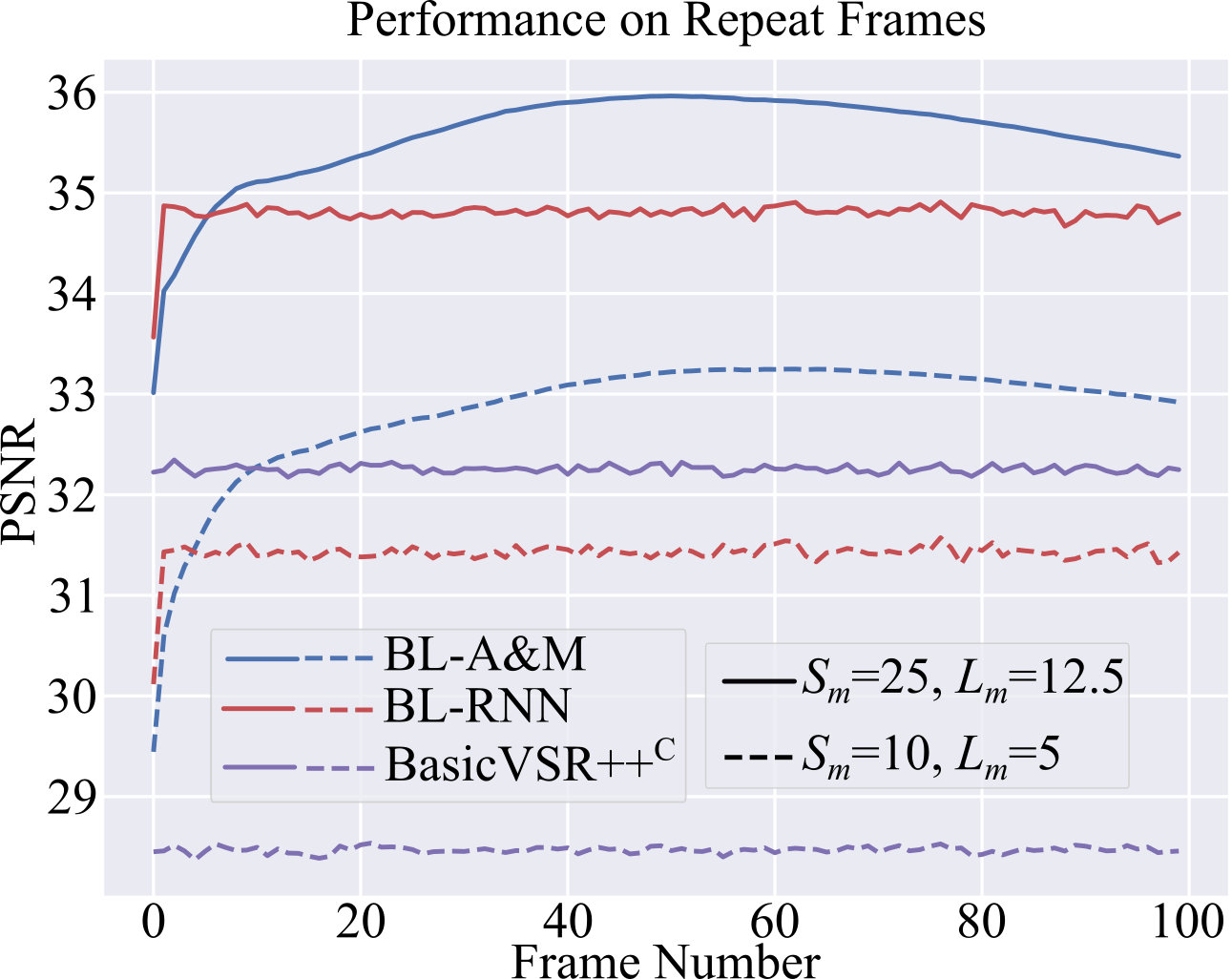}
\caption{\textbf{Repeated Frames:} In this experiment, we copy the first frame of each test video 100 times to simulate a scene without motion. We find RNN-BL and BasicVSR++$^C$ have constant performance after the first frame, whereas BL-A$\&$M improves for the first 50 frames. After 50 frames the BL-A$\&$M model is outside of its training environment so has a slight decrease in performance.
  \label{fig:stationary_experiment}}
\end{figure}

\subsection{Use of Temporal Information}
In this experiment, we test the networks abilities to use temporal information in the absence of motion. We conduct this experiment by simulating a no motion scenario by duplicating the first frame of each test video 100 times to create a video with 100 identical frames. We add simulated noise to these videos following our noise model for the case of $S_{m} = 25, L_{m} = 12.5$ and $S_{m} = 10, L_{m} = 5$. We test the performance of BL-A$\&$M, RNN-BL, and BasicVSR++$^C$ to see how well each is able to use temporal information. Because the frames have no motion the BL-A$\&$M model will average the prior frames together as input into the downstream network so will be forced to use the temporal information whereas the other models may not learn to use temporal information to its full extent. We plot the average test video performance by frame number in Fig.~\ref{fig:stationary_experiment}. 

We find that the recurrent models, RNN-BL and BasicVSR++$^C$, have constant performance across all time points except in the first frame for BL-RNN. This implies these recurrent models do not fully exploit temporal information in the case of no motion. The BL-A$\&$M improves sharply for the first 50 frames before slightly decreasing in performance. The slight decrease in performance is likely due to the fact that the BL-A$\&$M model was not trained with long repeated frame videos so these longer averages are outside of the training environment leading to a performance drop once far outside of examples in the training set. As expected temporal information provides a larger performance improvement for BL-A$\&$M at lower signal levels emphasizing that as models are pushed to work with less noise they will need to make better use of temporal information. However, it seems that current recurrent structures do not necessarily learn to use this information to its full extent but the performance improvement of BL-A$\&$M.

Interestingly, the BL-A$\&$M model also does not appear to be optimally using the temporal information. It is forced to average temporally, so after 3 frames at the $S_m=10$ noise level the average should have similar shot noise characteristics as the $S_m=25$ noise level case, only differing in the read noise. However, the performance at the $S_m=10$ three frames average is more than 2 PSNR below the $S_m=25$ first frame performance. This could be because the read noise spatial correlations are important for read noise removal and a simple average of the frames is not a sufficient statistic for this problem leading to reduced efficacy in read noise removal.  

\begin{figure*}[t!]
\centering
\includegraphics[width=0.7\linewidth]{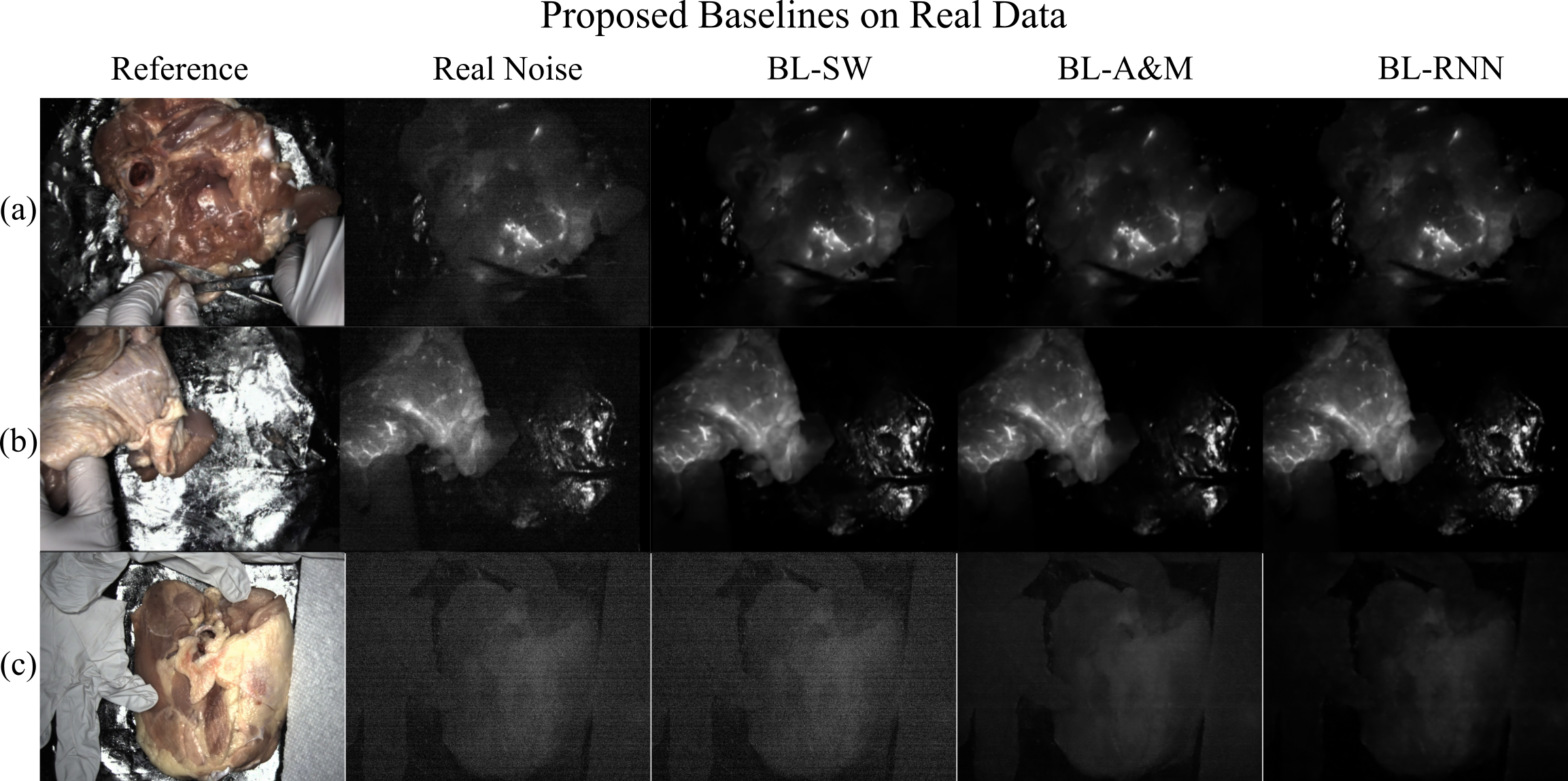}
\caption{\textbf{OL-2024 Results:} This figure shows the results of our baseline models (a-b) on 2 scenes from OL-Real. Notice how the hands in (a-b) are correctly removed by our baselines. (c) shows a scene with no fluorescent agent injected, BL-RNN is able to correctly remove most of the LLL in this scenario.
  \label{fig:real_results}}
\end{figure*}

\subsection{Real Test Data}
We test our models on OL-Real; however, because we lack ground truth we can not compute metrics on the results. Example result images are shown in Fig.~\ref{fig:real_results}(a-b). We find that all our baseline models are able to perform reasonably well on OL-Real. See supplement for video results.

In another test example we input real noisy data that contains no added fluorescence; the results are shown in Fig~\ref{fig:real_results}(c). For this test, we want a model that correctly finds that no fluorescence is present and outputs a image of all zeros. Note this training example is outside of the training regime of the models so unsurprisingly many of them struggle with this task. Surprisingly, BL-RNN is able to generalize and properly remove most of the LLL present in this image and is the only model that correctly removes the entire LLL coming from the gloves and paper towel on the sides of this scene.


\section{Future Outlook}

\noindent\textbf{Real Time: } In this work we did not focus on the real time performance characteristics of these algorithms, which will be a key part of future work. An extensive look at real-time options will need to consider many strategies including: caching, network pruning, half-point precision, image down-sampling, and other hardware optimizations. 

\noindent\textbf{Clinical Data: } Real clinical data will be a key part of deploying a real clinical system; however, capturing this data is expensive and time consuming so ensuring the correct data is collected is essential. Specifically, it will be important to be able to work with noisy data because ground truth bright fluorescence will not be available. While Noise2Noise~\cite{lehtinen2018noise2noise} and other variants~\cite{krull2019noise2void,pmlr-v97-batson19a} offer promising solutions to training with only noisy data, these solutions assume that the noise is zero-mean and LLL breaks this assumption. An important area of future work will be dealing with LLL with a lack of ground truth data; for example, by capturing multiple datasets focused on the LLL and fluorescence separately. Another strategy could try to exploit LLL and RV correlations to find LLL from LLL corrupted FV. We hope this paper can provide a starting point for this future work. 

\noindent\textbf{Evaluation: } Evaluation of different methods is key to understanding performance; however, most current cost functions are developed for natural images. A FGS specific lost function based on surgeons perception will be important for maximizing the value of FGS denoisers. Additionally, tying clinical outcomes to the evaluation will ensure useful performance. 

\noindent\textbf{Hardware Optimization: } Our work also sets the stage for better posed cost functions in hardware optimization in FGS. Currently, when selecting filters and excitation light sources a number of factors are considered including the tradeoff between LLL and fluorescence yield. It is possible to reduce LLL, but this may also lower the number of fluorescence photons captured by the FV camera. Understanding the exact cost function associated with this tradeoff space is difficult and currently an open problem. By using machine learning algorithm performance over a large simulated noise space it is possible to generate a cost landscape over $S_m$ and $L_m$ based on an objective function (i.e. LPIPS). This cost landscape could be used to rank candidate hardware configurations.

\section{Conclusion}
In this work, we considered the difficult problem of FGS video denoising and how it differs from conventional video denoising. We developed a realistic noise model that includes complicated read noise, and we proposed a solution for simulating LLL based on the RV. We explored how the unique FGS setting causes SOTA methods to struggle and proposed strong baseline models to guide future algorithmic development. We also included a number of evaluation settings for model comparison and evaluation. We hope this work sets the stage for future FGS denoising methods.

\small{\noindent\textbf{Acknowledgements: } This material is based upon work supported by the National Science Foundation (GRFP
DGE-1747503, CAREER 1846884), and the Wisconsin Alumni Research Foundation. The authors would like to also thank OnLume Inc. for access to the imaging system used for this work. }
{
    \small
    \bibliographystyle{IEEEtran}
    \bibliography{main}
}

\begin{IEEEbiography} [{\includegraphics[width=1in,height=1.25in,clip,keepaspectratio]{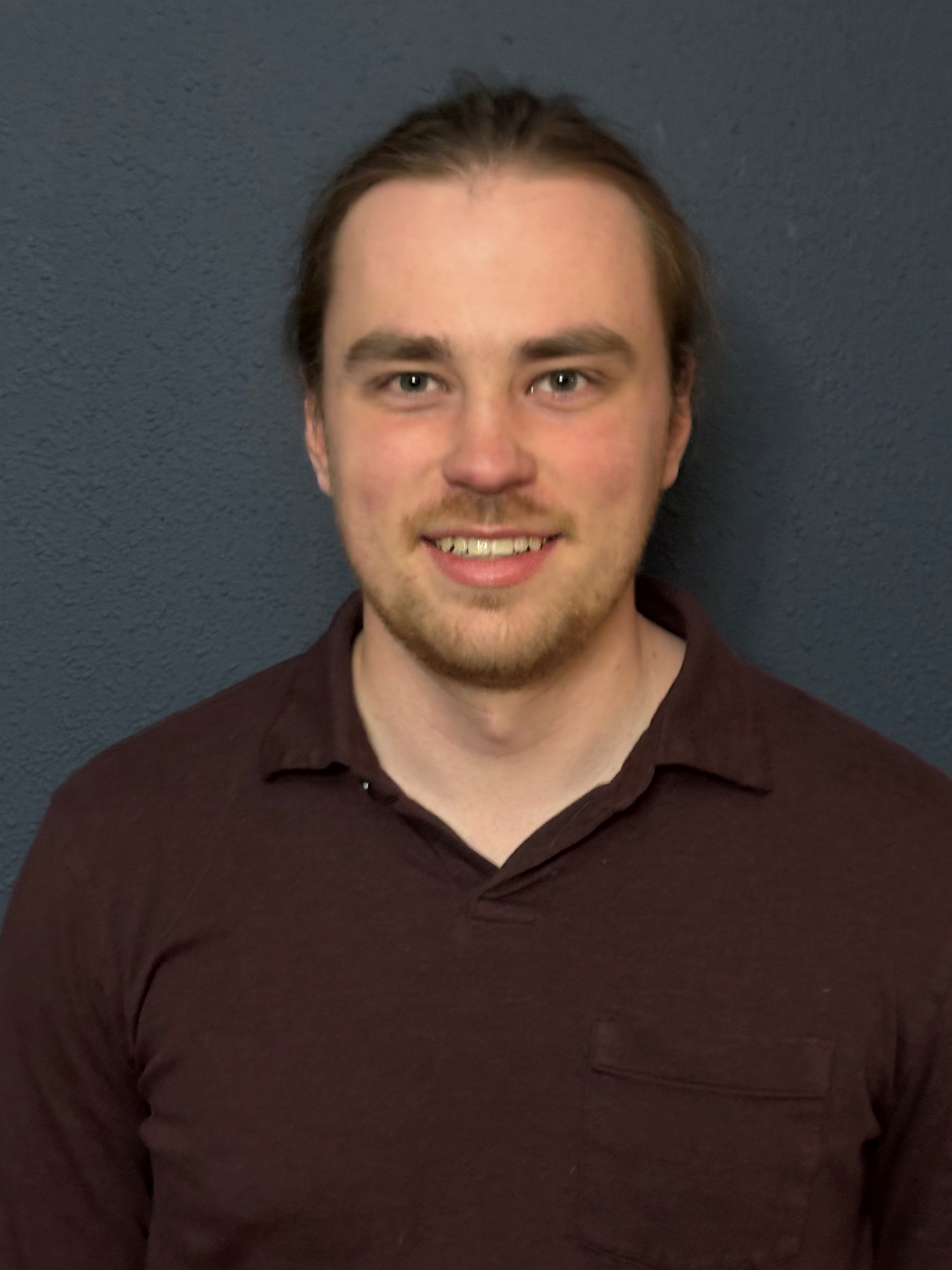}}]{Trevor Seets}
received his B.S. and M.S. degree in electrical engineering from the University of Wisconsin-Madison in 2019 and 2023, respectively. He is now a Ph.D. student under Professor Andreas Velten at UW-Madison. His research interests include computational optics, statistical signal processing, and imaging.
\end{IEEEbiography}

\begin{IEEEbiography} [{\includegraphics[width=1in,height=1.25in,clip,keepaspectratio]{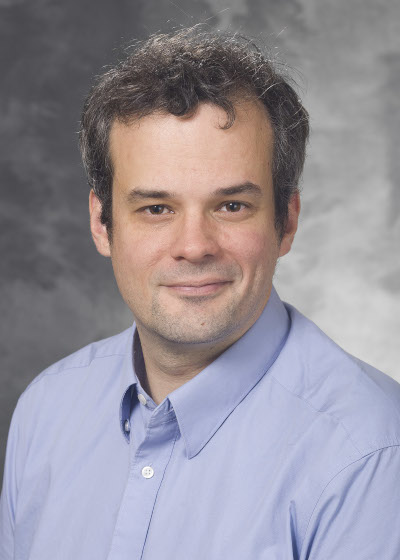}}]{Andreas Velten}
received the B.A. degree in physics from the Julius Maximilian University of Wurzburg, Wurzburg, Germany, in 2003, and the Ph.D. degree in physics from the University of New Mexico, Albuquerque, NM, in 2009. From 2010 to 2012, he was a postdoctoral research associate with the Massachusetts Institute of Technology, Cambridge, MA, USA, where he was working on high speed imaging systems that can look around
a corner using scattered light. From 2013 to 2016, he was an associate scientist with the Laboratory for Optical and Computational Instrumentation, University of Wisconsin–-Madison, Madison WI, USA, working in optics, computational imaging, and medical devices. Since 2016, he has been an associate professor with the Biostatistics and Medical Informatics, Electrical and Computer Engineering Department, University of Wisconsin–-Madison. His research focuses on performing multidisciplinary work in applied computational optics and imaging.
\end{IEEEbiography}

\newpage

\clearpage
\setcounter{page}{1}
\setcounter{figure}{0}
\setcounter{section}{0}
\setcounter{equation}{0}
\onecolumn
\normalsize
\begin{center}
\Large Appendix Document for\\[0.2cm]
\Large ``Video Denoising in Fluorescence Guided Surgery'' \\[1.5cm]
\end{center}

\begin{figure*}[b]
\centering
\includegraphics[width=0.9\linewidth]{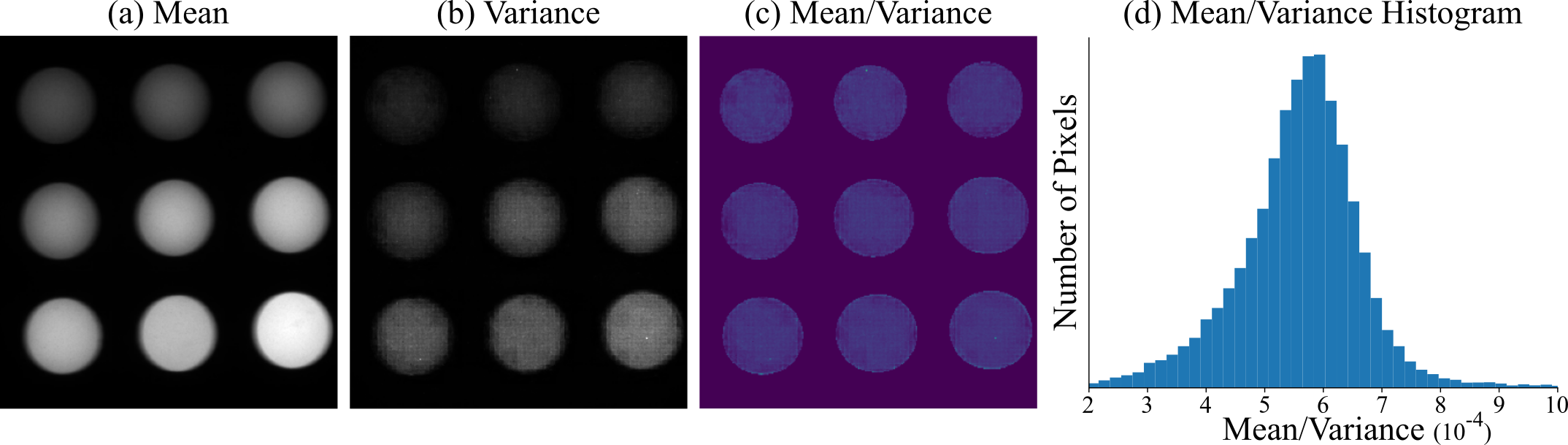}
\caption[Gain Calibration Images]{\textbf{Gain Calibration Images:} This figure shows the (a) mean, (b) variance, and (c) the mean divided by the variance of the Quel calibration phantom from OL-Phantom used to calibrate for our $K$ parameter. (d) shows the histogram of mean divided by variance values for the 9 wells. The mean value of this histogram is our calibrated $K$ value. Notice that the mean divided by variance values are similar in all wells indicating that a Poisson distribution is correct for the brightness level of this phantom.
  \label{supp_fig:quel_phantoms}}
\end{figure*}

\section{Calibration and Data Generation Details}

\subsection{Gain (K)}
In order to solve for the $K$ in our noise generation model in Eq.~\ref{eq:noise_model}, we do the following calibration procedure. First, we capture a video with 1,830 frames of the Quel calibration phantom \cite{ruiz2020indocyanine} (OL-Phantom). The Quel phantom contains 9 wells of varying ICG mimicking fluorescence concentrations. We then take the mean and variance of each pixel across time. We then assume that the Poisson noise term dominates (i.e. read noise is negligible) so our measurements, $I$, will be Poisson random variables with mean equal to variance. Therefore, $K$ can be estimated as,

\begin{align}
    I &= K\text{Pois}(S) \\
    \text{Var}[I] &=K^2S  \\
     \text{E}[I]&= KS\\
    K &= \frac{\text{Var}[I]}{\text{E}[I]} \label{supp_eq:K_est}
\end{align}

where E is the expectation across time, and Var is the variance across time. We use Eq.~\ref{supp_eq:K_est} to estimate a $K$ value for each pixel in a Quel phantom well, we average the estimates to give a final estimate of $K=\frac{1}{1764}$ when $I$ is scaled between 0 and 1.

\subsection{Quantizer}
The OnLume Avata System by default captures 12-bit images and exports 8-bit mp4s. In training, we use a quantizer equivalent to the internal 12-bit quantization to match performance as if the models are run on the system. The quantizer function, Quant, rounds measurements to the nearest multiple of $\frac{1}{2^{12}}$ and clips the values between $[0,1]$.

\subsection{Read Noise Scale}
The OnLume Avata System by default captures 12 bit images and exports 8 bit mp4s. In order to capture higher bit-depth examples of the read noise, we use the digital gain parameter of the system. Therefore, our read noise samples use a different quantizer than the quantizer used in calibration of the $K$ parameter. We match the scales of the read noise and shot noise terms by dividing the sampled read noise frames by $R_{m}$, we sample $R_{m}$ values between $[4,8]$ during training and use $R_{m} =6.0$ during testing which is calibrated so our scaled read noise frames match the digital gain used to calibrate $K$. 

\subsection{Data Augmentation}
In order to augment our data and not over fit to a given set of camera parameters we randomize different camera parameters over a small range, while testing with our calibrated parameters. Table~\ref{supp_tab:camera_params} shows the camera values  used in training and testing.

\begin{table}[t]
    \centering
    \begin{tabular}{c|c|c|c}
         Param & Min & Max & Test  \\
          $\frac{1}{K}$ &  1200 &  2400  & 1763.5\\
          $R_{m}$ & 4 & 8 & 6\\          $S_{m}$ & 10 & $\frac{1}{2K}$  & Varies\\
          $L_{m}$ & 0 &  $S_{m}$ & Varies
    \end{tabular}
    \caption[Noise Simulation Parameters]{This table shows the camera values used in training and testing.}
    \label{supp_tab:camera_params}
\end{table}

\subsection{LLL-PN Details \label{suppl_sec:lll_pn}}
\textbf{Network and Training Details: } For our LLL-PN, $f_{\text{LLL-PN}}(I ; \theta)$, we use a lightweight version of NafNet~\cite{chen2022simple}. We use the OL-LLL dataset for training which contains video of a mock chicken thigh surgery without fluorescent dyes, therefore the fluorescent frames only contain the LLL term ($L^{LL}$). We train our LLL-PN parameters, $\theta$, with the following loss function,

\begin{equation}
    l_p(I_{\text{ref}},I_{\text{FL}},\theta) = || f_{\text{LLL-PN}}(I_{\text{ref}} ; \theta) - I_{\text{FL}}||_p
\end{equation}

where $I_{\text{ref}}$ is a reference frame, $I_{\text{FL}}$ is a fluorescent frame, and $||\cdot||_p$ is the $l_p$-norm we try $p=1,2$. We train our network with the Adam optimizer \cite{Kingma2014AdamAM} ($\beta_1=0.9, \beta_2 =0.999$) with initial learning rate of $10^{-4}$ and batch size of 10 for 100 epochs with a cosine annealing learning rate decay. $I_{ref}$ and $I_{FL}$ are the reference and fluorescence frames respectively.

\begin{figure*}[t]
\centering
\includegraphics[width=1.0\linewidth]{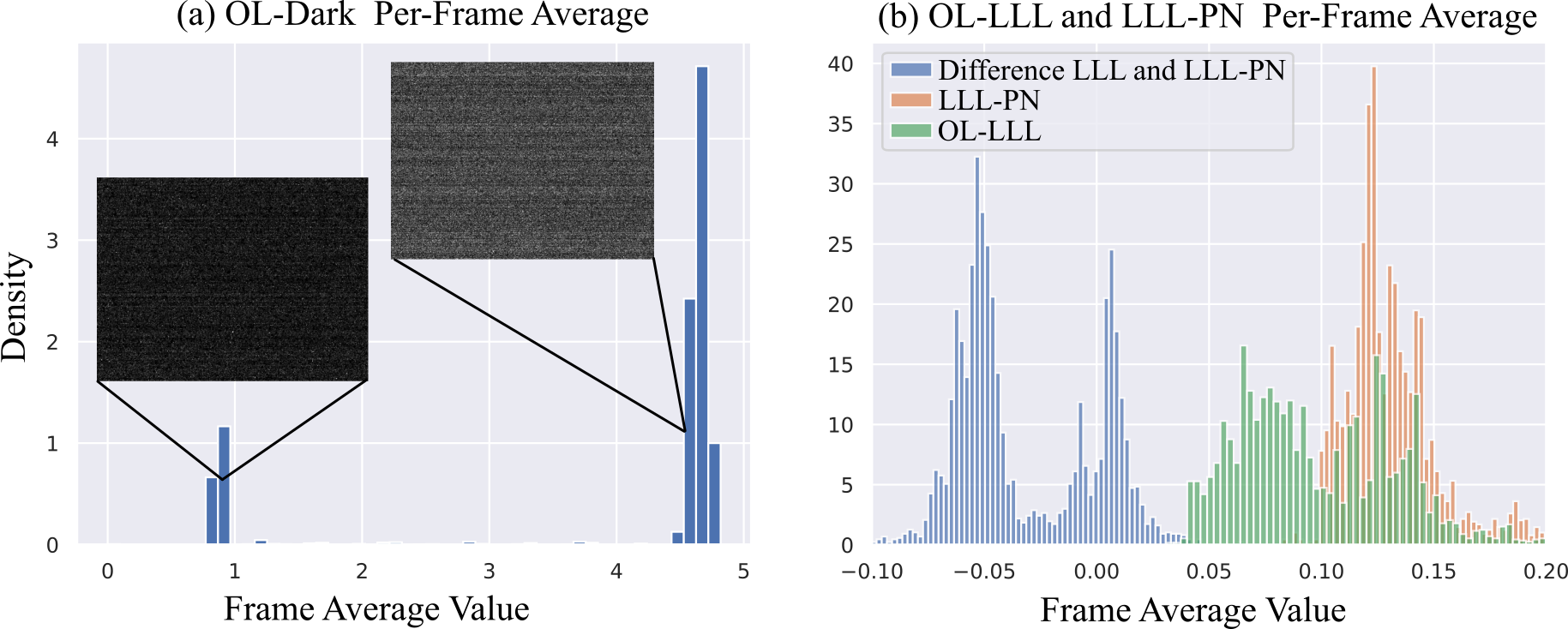}
\caption[Flicker Noise]{\textbf{Flicker Noise:} (a) shows the histogram of the per-frame average values of the OL-Dark dataset. The read noise pattern of our camera sensor exhibits strong flicker noise leading to a bimodal distribution, one example from each peak is shown. (b) shows the per-frame average of the OL-LLL dataset (green), the LLL-PN (orange), and the difference between the two (blue). Due to the bimodal flicker noise our LLL-PN predicts the "high" mode of the flicker; however, we are able to obtain the bimodal distribution when subtracting out our background indicating strong performance with subtracting out the LLL and leaving the read noise.  
  \label{suppl_fig:flicker}}
\end{figure*}

\noindent\textbf{Choice of loss: } We try both the $l_1$ and $l_2$ loss to train our LLL-PN. Because we are training in the Noise2Noise paradigm, it is important to consider the sources of noise in our training data. Because our read noise exhibits a strong flicker noise that is non-zero mean, we expect the $l_1$ loss to work better as it is a median seeking loss whereas $l_2$ is an average seeking loss \cite{lehtinen2018noise2noise} and will predict an average of the flicker which is not dependent on the LLL. We evaluate $l_1$ and $l_2$ errors of the networks trained using both training losses; the results are shown in Table~\ref{supp_tab:LLL-PN}. We find that the $l_1$ training loss provides better performance so we use this network as our final LLL-PN. Note that the calculated error metrics will include noise energy from read noise and Poisson noise from the background, so the minimum error will never reach 0 as our LLL-PN can only remove the bias term associated with the background. 

\noindent\textbf{Flicker noise: }We find the median seeking $L_1$ loss performs better because our read noise exhibits strong flicker noise producing a bimodal noise distribution with different additive bias terms, shown in Fig.~\ref{suppl_fig:flicker}(a). By using an $L_1$ loss, the LLL-PN predicts the additive bias term that is most present in the training data (the brighter flicker). Fig.~\ref{suppl_fig:flicker} shows histograms of the OL-LLL test set per-frame average image values for the noisy LLL, $f_{LLL}(R^v_t)$, and the difference between the predicted and noisy values. Notice that, as expected, the bimodal distribution of the flicker noise is much more visible in the difference histogram; although, the bimodal distributions are wider which can be accounted for by shot noise and any mistakes made by the LLL-PN.

\begin{table*}[h]
    \centering
    \begin{tabular}{||c|c|c|| c ||}
    \hline
          \hline
          Training Loss & $l_1$ & $l_2$ & Fluorescent Frame Norms\\
          \hline
          Errors & 0.0621/0.0762&0.0643/0.0785 & 0.100/0.127 \\
          \hline
    \end{tabular}
    \caption[LLL-PN loss function comparison]{\textbf{LLL-PN loss function comparison:} we compare the errors ($l_1/l_2)$ of our LLL-PN for different loss functions. We also include the noisy fluorescent frame $l_1$ and $l_2$ norms to show how much background signal is subtracted.}
    \label{supp_tab:LLL-PN}
\end{table*}

\newpage

\section{Training Parameters: }
The values we use for training all models are shown in Suppl. Tab.~\ref{tab:training_params}, full config files will be available with the code. One key challenge in video denoising is balancing training time with the number of parameters. We find this is even more true in FGS denoising because of the large noise parameter space that needs to be sweeped across requiring many parameters to deal with this large number of potential noise levels. Our BL-RNN model is able to train quickly to convergence while maintaining a large number of parameters which is not true for other models such as BasicVSR++$^C$. 

\begin{table*}[h]
    \centering
    \caption[Model Training Details]{\textbf{Training Details: } In each epoch we go through every 100 frame non-overlapping partition (768 sets) in the training set, for each 100-frame set in a batch we pull out Num T frames for the full batch used in one iteration of training. Total time is the amount of hours it takes to train the model with a single A100 GPU. Iterations is the number of gradient decent steps done in total, and total seen is the number of total frames seen during training. BL-SW and FastDVDnet are trained using 5 frames from every video but only denoise the last frame of this group. To train Bl-A$\&$M we first train the NafNet32 backbone, the values after the $+$ sign in the table indicate the where the pre-training parameters of NafNet in Bl-A$\&$M for differ from the NafNet32 full training.
    \label{tab:training_params}}
    \normalsize{
    \begin{tabular}{c|c|c|c|c|c|c|c|c|}
           Model &Seconds/epoch & Epochs & Num T & Batch &Total time (hr) & Iterations & Images Seen  \\
          \hline
          BL-RNN & 90 & 3k & 4 & 15 & 75  & 153k & 9.2M \\
          BL-SW & 90 & 6k & 5 (1) & 14 & 150  & 324k & 22.7M \\
          Bl-A$\&$M & 800& 1.2k+1.6k & 5 & 8 & 267+69 & 111k+302k &  4.5M+12.1M \\
          NafNet32 & 155 & 3k & 10 & 4 & 129 & 567k & 22.7M  \\
          BasicVSR++$^C$& 1100 & 1000 & 30 & 2 & 306 & 377k & 22.7M \\
          FastDVDnet$^C$ & 90 & 6k & 5 (1) & 14 & 150  & 324k & 22.7M \\
          OFDVDnet$^C$ & 950 & 1.2k & 6 & 8 & 316.67 & 114k &  5.4M  \\
    \end{tabular}
    }
\end{table*}
\newpage
\newpage
\section{Causal BasicVSR++ on Davis\label{suppl:davis_results}}

We retrain our modified causal BasicVSR++ model on the conventional video denoising Davis dataset following the training scheme from the BasicVSR++ paper~\cite{chan2022basicvsr++}. We also train NafNet32 and NafNet64 on this dataset to compare the results with our choice of image denoiser baseline. Our quantitative results are summarized in Supp. Table.~\ref{supp_tab:davis_results}. We find that converting BasicVSR++ to a causal network greatly hurts the performance at large noise levels.

\begin{table*}[]
\centering
\begin{tabular}{c|c|c|c|c|c|c}
& VBM4D & FastDVDnet &BasicVSR++$_2$ &Causal BasicVSR++$_2$&NafNet32 &NafNet64 \\
\hline
$\sigma=10$ & 37.58/- &38.71/0.9672 & 40.97/0.9786 &39.57/0.9721& 36.32/0.9455& 38.30/0.9623 \\
$\sigma=20$ & 33.88/- &35.77/0.9405&38.58/0.966 &36.72/ 0.9516& 33.14/0.8998& 34.86/0.9267\\
$\sigma=30$ & 31.65/- & 34.04/0.9167&37.14/0.9560 &35.02/0.9328& 31.19/0.8547&32.96/0.8957\\
$\sigma=40$ & 30.05/- & 32.82/0.8949&36.06/0.9459& 33.76/0.9143& 29.76/0.8081&31.67/0.8689\\
$\sigma=50$ & 28.8/-& 31.86/0.8747&35.18/0.9358 &32.74/0.8951& 28.55/0.7545&30.70/0.8460\\
\hline
Average & 32.39/- & 34.64/0.9188 & 37.58/0.9566 &35.56/0.9332& 31.79/0.8526&33.67/0.8999
\end{tabular}
\caption{NafNet and Causal BasicVSR++ on Davis:} Here we report the standard quantitative results (PSNR/SSIM) for video denoising on the Davis dataset for our trained models: Causal BasicVSR++$_2$, NafNet32 and NafNet64.
\label{supp_tab:davis_results}
\end{table*}

\newpage
\newpage
\section{Training Instability of Recurrent Candidate Models}
\begin{figure*}[t!]
\centering
\includegraphics[width=0.9\linewidth]{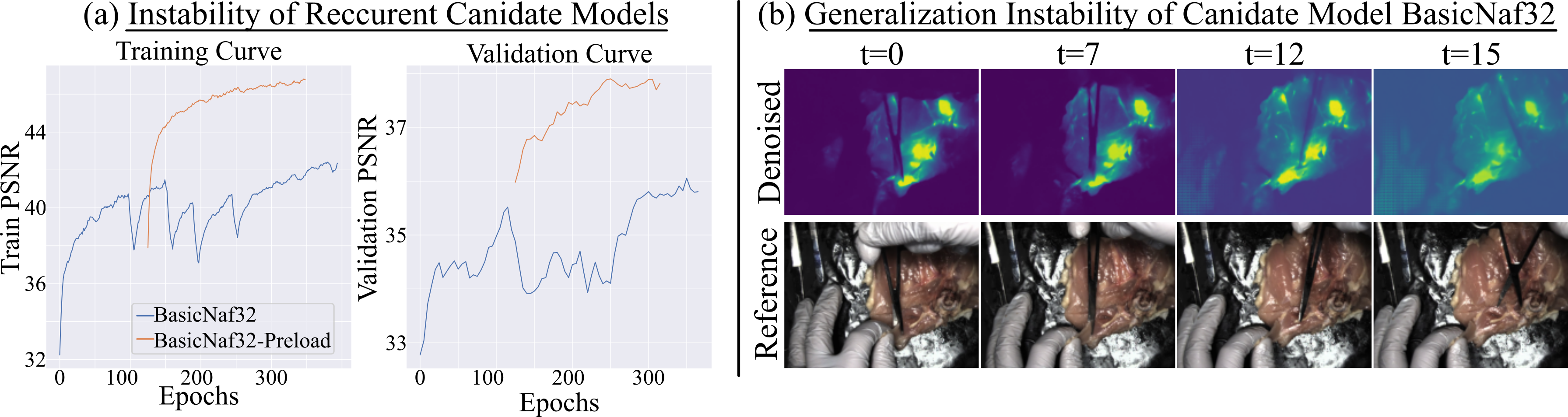}
\caption{\textbf{Instability of Recurrent Candidate Models:} (a) Training and validation curves for the BasicNaf32 and BasicNaf32-Preload candidate models. The BasicNaf32 model exhibits training instability after 100 epochs of training (60 hours). While our BasicNaf32-Preload exhibits smoother training. Note BasicNaf32-Preload curves are offset by the pre-loaded image only model training time in equivalent BasicNaf32 epochs. (b) BasicNaf32-Preload has trouble generalizing to the full resolution test set that results in artifacts for later recurrent frames. 
  \label{supp_fig:basicnaf32_v3_unstable}}
\end{figure*}

We found that training recurrent networks can lead to instability even late in the training process after many hours of training. This makes fast iteration of recurrent structures challenging because time is wasted on tuning the learning rate or other hyper parameters. One of our candidate models was a combination of NafNet and BasicVSR++ called BasicNaf32, which uses the  Unet structure of NafNet but the recurrent temporal feature propagation of BasicVSR++. We found that one hyperparameter combination of BasicNaf32  becomes unstable after about 60 hours of training, the training curve is shown in Fig.~\ref{supp_fig:basicnaf32_v3_unstable}(a). A preloading strategy is used to train BasicNaf32-Preload: this strategy first trains the network for 1600 epochs on only images and provides stability to the training process. However, a surprising generalization problem occurs in both models where they are unable to generalize from the trained 256 by 256 images to the full sized 768 by 1024 test images. This instability occurs after about 10 frames in this recurrent structure as shown in Fig.~\ref{supp_fig:basicnaf32_v3_unstable}(b). We found this unstable behavior to occur often within combinations of NafNet and BasicVSR++ because of this finding we chose to use a simpler recurrent structure for our BL-RNN.

\newpage
\section{No Reference Frame Qualitative Result}

\begin{figure*}[t!]
\centering
\includegraphics[width=0.9\linewidth]{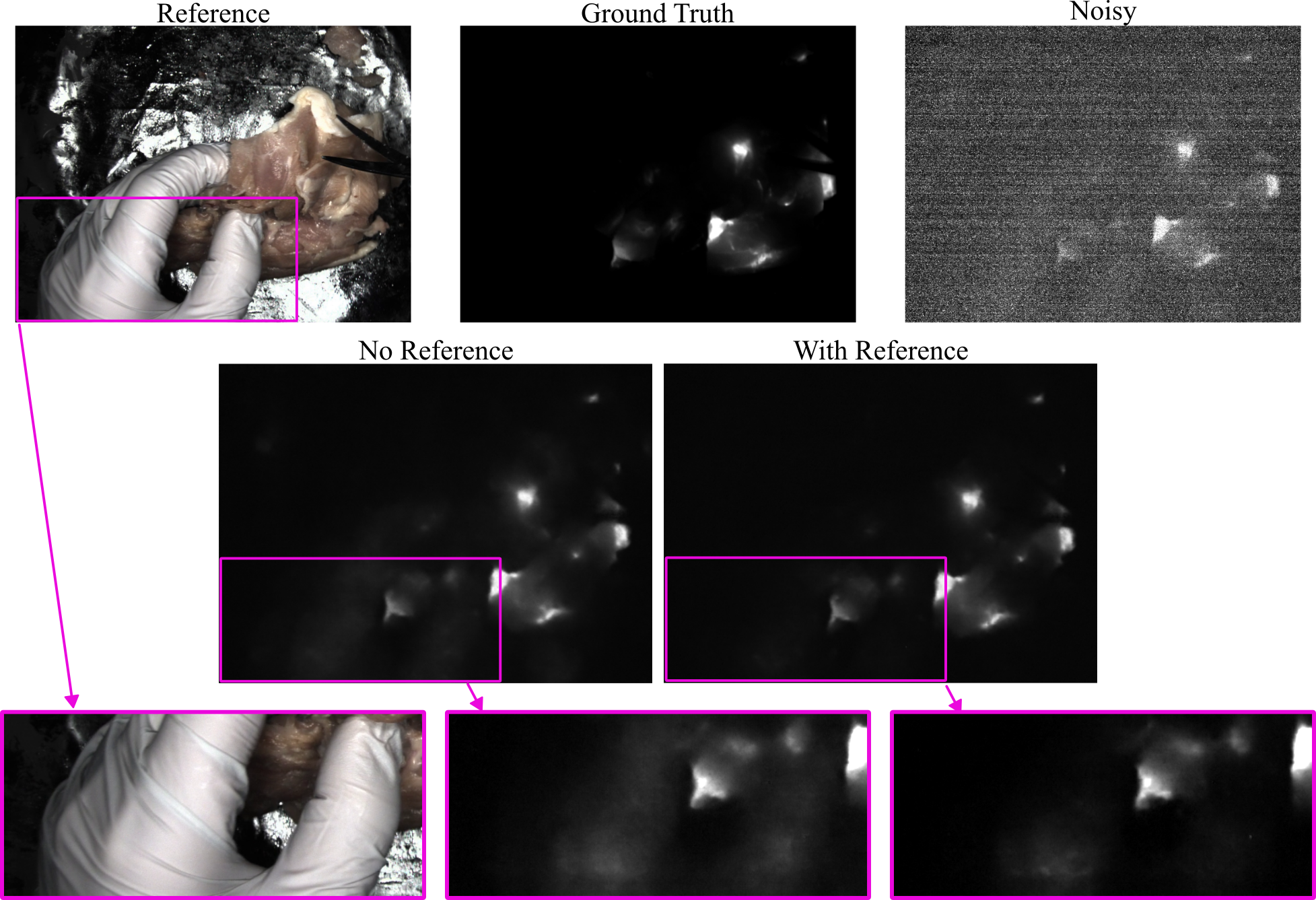}
\caption{\textbf{No Reference: }This figure shows a case where the reference frame is key in the denoising process to properly remove the LLL associated with a hand. The noise parameters used for this example are $S_m=L_m=25$. \label{supp_fig:no_ref}}
\end{figure*}

 We find that the RV is key in removing LLL; an example is shown in Suppl. Fig~\ref{supp_fig:no_ref}. In this example, the surgeons hand is clearly visible in the BL-RNN network that does not have access to the reference frames whereas the BL-RNN network that uses the reference frames is able to better remove the hand.

 \section{Video Results}
Trimmed (due to file size limitation) video results are available in the supplement. 
\begin{enumerate}
    \item "real\_short.mp4" : baseline results on OL-Real
    \item "Short\_sm25\_lm\_25.mp4" : baseline results on $L_m=25=S_m$ simulated data
    \item "Short\_sm50\_lm\_50.mp4" : all model results on $L_m=50=S_m$ simulated data. In general the baseline models obtain better temporal consistency than either NafNet32 or BasicVSR++$^C$.
\end{enumerate}

Full video results are available at \cite{ol24_dataset}. Reviewer link to dataset with video results: \href{https://datadryad.org/stash/share/k7lt5eto1DvreodhSqgpArB1vHYECchBzJk1JARvunU}{https://datadryad.org/stash/share/k7lt5eto1DvreodhSqgpArB1vHYECchBzJk1JARvunU}

\vfill

\end{document}